\documentclass[lettersize,journal]{IEEEtran}
\usepackage{amsmath,amsfonts}
\usepackage{multirow}
\usepackage{subfigure}
\usepackage[linesnumbered,ruled,vlined,onelanguage]{algorithm2e}
\usepackage{color}
\usepackage{array}
\usepackage[caption=false,font=normalsize,labelfont=sf,textfont=sf]{subfig}
\usepackage{textcomp}
\usepackage{stfloats}
\usepackage{url}
\usepackage{verbatim}
\usepackage{graphicx}
\usepackage{cite}
\begin{document}

%\title{Improve pixel cohesion and focus on environment for text detection}
\title{Focus Entirety and Perceive Environment for Arbitrary-Shaped Text Detection}
\author{
	Xu~Han,  
	Junyu~Gao,~\IEEEmembership{Member,~IEEE,}
	Chuang~Yang, 
	Yuan~Yuan,~\IEEEmembership{~Senior Member,~IEEE}
	
	and Qi~Wang,~\IEEEmembership{~Senior Member,~IEEE}
	% <-this % stops a space
	\thanks{
		
		X. Han, C. Yang are with the School of Computer Science, and with the School of Artificial Intelligence, Optics and Electronics (iOPEN), Northwestern Polytechnical University, Xi'an {\rm 710072}, P. R. China. (E-mail: hxu04100@gmail.com, omtcyang@gmail.com).
		
		%			J. Gao is with the School of Artificial Intelligence, Optics and Electronics (iOPEN), Northwestern Polytechnical University, Xi'an {\rm 710072}, P. R. China. E-mail: gjy3035@gmail.com.
		%		
		%			Y. Yuan is with the School of Artificial Intelligence, Optics and Electronics (iOPEN), Northwestern Polytechnical University, Xi'an {\rm 710072}, P. R. China. E-mail: y.yuan1.ieee@gmail.com.
		%		
		%			Q. Wang is with the School of Artificial Intelligence, Optics and Electronics (iOPEN), Northwestern Polytechnical University, Xi'an {\rm 710072}, P. R. China. E-mail: crabwq@gmail.com.
		
	}

	\thanks{J. Gao, Y. Yuan, and Q. Wang are with the School of Artificial Intelligence, Optics and Electronics (iOPEN), Northwestern Polytechnical University, Xi'an {\rm 710072},  P. R. China. (E-mail: gjy3035@gmail.com, y.yuan1.ieee@gmail.com, crabwq@gmail.com). 
		}
\thanks{This work was supported by the National Natural Science Foundation of China under Grant U21B2041.
}%
\thanks{Qi Wang is the corresponding author.}
}

% <-this % stops a space
%\thanks{Manuscript received April 19, 2005; revised January 11, 2007.}}

\markboth{{IEEE} Transactions on MULTIMEDIA}%
{Shell \MakeLowercase{\textit{et al.}}: A Sample Article Using IEEEtran.cls for IEEE Journals}


\maketitle

\begin{abstract}
Due to the diversity of scene text in aspects such as font, color, shape, and size, accurately and efficiently detecting text is still a formidable challenge. 
Among the various detection approaches, segmentation-based approaches have emerged as prominent contenders owing to their flexible pixel-level predictions. 
However, these methods typically model text instances in a bottom-up manner, which is highly susceptible to noise. In addition, the prediction of pixels is isolated without introducing pixel-feature interaction, which also influences the detection performance.
To alleviate these problems, we propose a multi-information level arbitrary-shaped text detector consisting of a focus entirety module (FEM) and a perceive environment module (PEM). The former extracts instance-level features and adopts a top-down scheme to model texts to reduce the influence of noises.
Specifically, it assigns consistent entirety information to pixels within the same instance to improve their cohesion. In addition,  it emphasizes the scale information, enabling the model to distinguish varying scale texts effectively.
The latter extracts region-level information and encourages the model to focus on the distribution of positive samples in the vicinity of a pixel, which perceives environment information. It treats the kernel pixels as positive samples and helps the model differentiate text and kernel features.
Extensive experiments demonstrate the FEM's ability to efficiently support the model in handling different scale texts and confirm the PEM can assist in perceiving pixels more accurately by focusing on pixel vicinities. Comparisons show the proposed model outperforms existing state-of-the-art approaches on four public datasets.

\end{abstract}

\indent

\section{Introduction}
\IEEEPARstart{O}{ver} the past few years, research on scene text detection has gained increased concerns due to its various applications, including license plate detection, signboard reading, autonomous driving, and scene understanding. With the rapid development of object detection and image segmentation,  scene text detection \cite{ad, rmstd,9103135,leaftext, astd, wang2020r, zoom,han2023text} achieves significant progress. However, accurately locating scene text remains tricky due to font, color, and scale variation. The irregular shape is still the most formidable challenge of them.
\begin{figure}
	\centering
	\includegraphics[width=0.98\linewidth]{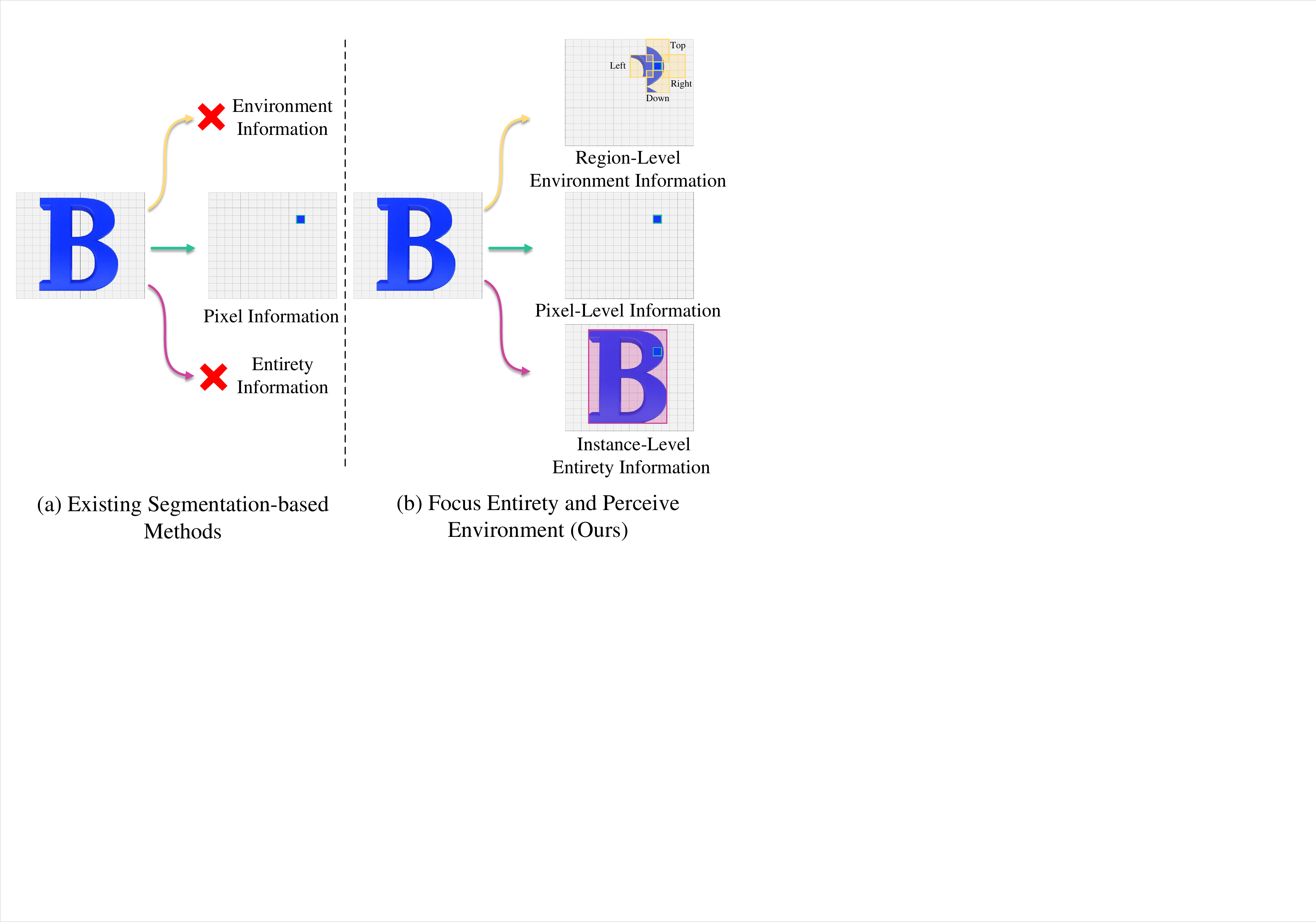}
	
	\caption{ Illustration of the multi-level information extraction for existing segmentation-based methods and ours. (a) Existing segmentation-based methods \cite{pse}, \cite{db}, \cite{db++} only focus on pixel-level information. (b)  Our method further extracts region-level and instance-level features to suppress the noise. }
	\label{first2}
\end{figure}

Among the numerous recent advanced approaches, segmentation-based approaches stand out as their flexible pixel prediction can cope with it.
PSENet \cite{pse} produces multi-scale pixel-level predictions to identify different scales of kernels. DBNet \cite{db} predicts the threshold and whether the pixel is a kernel or not. Although the above methods inherit and develop the advantage of pixel prediction, they still model text instances in a bottom-up manner, which is highly susceptible to noise. Moreover, these methods focus only on distinguishing text pixels from non-text pixels, ignoring essential features of the text instance. The fundamental purpose of text detection is to locate text instances.
To address the above problem, we propose a focus entirety module (FEM), which helps the proposed model extract instance-level features and utilizes a top-down scheme to model texts which reduces the influence of noises. It assigns consistent information to the pixels within the same instance to strengthen the cohesion of pixels.  In addition, the FEM encourages pixels to focus on the scale of instances and helps the model recognize features at different size instances to deal with text scale variations.
%Learning the characteristics variance of different text instances can enhance the cohesion of pixels that belong to the same instance.  
%Considering this, Bottom-up modeling approaches are susceptible to noise interference, and learning instance-level features can mitigate this issue. Furthermore, FEM helps the model recognize features at different size instances to deal with text scale variations.
%PSENet \cite{pse} makes multiple predictions for pixels to determine whether they belong to different scales of kernels. DBNet \cite{db} predicts the thresh of pixel while predicting it is kernel or not.   Although these method solve the problem of irregular shape to some extent, they only focus on whether the pixel is text or not and ignore some features of text instance which it belong to. As the essential goal of text detection is to distinguish text instances, learing the features of different text instance is benefical for understanding text properties that enhance the sense of belonging of pixels to the instance. The bottom-up modeling approach leads to model that are susceptible to noise interference, and learning  instance-level features could alleviate this issue. Consider this, we prepose a focus entirety module (FEM). It ensures that the model learns information about the pixel itself while forcing the model to focus on the overall features of the instance to which the pixel belongs. In addition, FEM assists the model to recognize features at different scales to cope with the challenge of large text scale variations.

Furthermore, existing segmentation-based methods focus only on predicting single pixels isolated without introducing information interaction. For example,
TextLeaf \cite{leaftext} focuses on the text kernel mask and rebuilds the instances by predicting the lateral and thin veins. CT-Net \cite{ct} predicts the kernel probability map and centripetal shift map to obtain detection results. 
However, it results in separate predictions of each pixel with no sufficient connection between them, influencing the detection performance.
It is considered that the visual system often utilizes the distribution of the surrounding environment to determine the properties of an object that is challenging to judge. The model should focus on information about individual pixels and the surrounding environment to help construct the entire textual knowledge system. 
To be specific, we propose a perceive environment module (PEM), which extracts region-level features and facilitates predicting peripheral pixel interactions to obtain synergistic progress. It perceives the positive sample distribution around the pixel in four directions to recognize hard-to-identify pixels effectively. Furthermore, the PEM treats the kernel pixels as positive samples and helps the model differentiate text and kernel features, improving the incomplete kernel semantics.
%All of these methods focus only on the prediction of a single pixel, and the predictions between each pixel are not sufficiently connected, making these predictions separate. Considering that the human visual system often assists in judging that target based on the distribution of the surrounding environment when it is difficult to judge the properties of that target. For example, if we observe a word "as?itant", where the "?" represents an unpredictable letter, we can determine the "?" in this case based on the surrounding environment is "s". 
%We believe that the model should not only focus on the information related to pixel itself, but also have an awareness of the surrounding environment to help the model construct the entire textual knowledge system well. Accordingly, we propose an environment-aware module (PEM) that perceives the positive sample distribution around the pixel in four directions to effecvtive recognize hard-to-identify pixels.

As we can see from Fig. \ref{first2}, existing segmentation-based methods \cite{pse, db, db++} model text instances in a bottom-up manner that only focus on pixel-level features lack coarse global features, which is susceptible to noise interference. The proposed method is named focus entirety and perceive environment (FEPE), which models text structure from three levels: coarse global features (instance-level), fine-grained local features (pixel-level), and their intermediate state (region-level).
Additionally, FEM and PEM can be removed during the testing phase. This means they improve the accuracy without affecting the inference speed and can be further integrated with other methods to improve their performance.
The main contributions of this work are as follows:
\begin{enumerate}
\item A focus entirety module (FEM) is proposed to extract instance-level features and model texts in a top-down scheme that reduces the influence of noise.
It assigns consistent entirety information to pixels within the same instance to improve their cohesion and emphasizes the scale of instances to which pixels belong, enabling the model to distinguish varying scale texts effectively.

\item A perceive environment module (PEM) is proposed to extract region-level information and encourage the model to focus on the distribution of positive samples in the vicinity of a pixel, which perceives environment information. It treats the kernel pixels as positive samples and helps the model differentiate text and kernel features, improving the incomplete kernel semantics.

\item  An efficient and effective text detector is proposed based on the above modules named FEPE, which attend simultaneously to coarse global features and fine-grained local features. It achieves state-of-the-art (SOTA) performance on multiple public benchmarks, which include numerous horizontal, rotated, and irregular-shaped texts.
\end{enumerate}

The rest of the paper is structured as follows. Some related work is presented in Section \ref{related}. Section \ref{method} describes the detail of FEM, PEM, and FEPE. Furthermore, we describe the multi-task loss used and the training details. In Section \ref{experiment}, ablation studies on four benchmarks strongly demonstrate the superiority of the proposed FEM and PEM. In addition, extensive experiment results are compared with state-of-the-art methods, proving the superiority and advancement of FEPE. Finally, the whole paper is summarized in Section \ref{conclusion}.
\section{Related work}
\label{related}
Deep learning has rapidly advanced in recent years, making significant progress in text detection. Existing methods are generally divided into regression-based methods, connected-component-based methods, and segmentation-based methods. The related works are briefly introduced as follows.

\subsection{Regression-based methods}
The majority of regression-based methods for text detection are inspired by object detection frameworks, such as Faster-RCNN \cite{ren2015faster}, and refined based on the characteristics of the text. Liao \emph{et al.}  proposed TextBoxes  \cite{liao2017textboxes}, which detect texts by revising the anchor and convolution kernels. Then, TextBoxes++ was \cite{liao2018++} proposed to cope with multi-directional text, which adds an angle parameter. Zhou \emph{et al.} \cite{east} divided text as rotated text and quadrangle text to predict different parameters based on FCN \cite{fcn}. Liao \emph{et al.} proposed RRD \cite{rrd}, which used rotation-sensitive features to detect oriented texts. He \emph{et al.} proposed SSTD \cite{sstd}, which used an attention module and an auxiliary loss to obtain detection results. Most of the above methods are limited by sophisticated post-processing, which affects their development. Moreover, the irregular-shaped text is a tricky problem for the above methods. Dai \emph{et al.} \cite{pcr} proposed progressive contour regression to cope with it, which iterative update text contours. The initial result is horizontal text, which is gradually optimized to multi-directional and irregular-shaped text. FCENet \cite{fcenet} and ABCNet \cite{abcnet} represented text contours by Fourier Signature Vector and Bezier Curve, respectively. Although the above methods can deal with irregular-shaped text, the complicated structure influences the efficiency.
\subsection{Connected-component-based methods}
Connected-component-based methods locate and group characters or parts of instances to reconstruct instances. CRAFT \cite{craft} modeled text instance by judging the proximity of the characters to each other. DRRG \cite{drrg} utilized graph convolutional networks (GCN) to infer the relationships between text parts. PixeLink \cite{pixellink} predicted pixel score map and the relationships with surrounding pixels to detect text. SegLink \cite{seglink} represented text instances as segments and links that merged segments according to the predictions of links. Long \emph{et al.}  proposed TextSnake\cite{textsnake}, which represents text like a snake. It utilized circles to describe text components. Although connected-component-based approaches work well when dealing with irregular-shaped texts, the complex merging process remains an open problem.
\begin{figure*}[t]
	\centering
	\includegraphics[width=0.98\linewidth,scale=1.0]{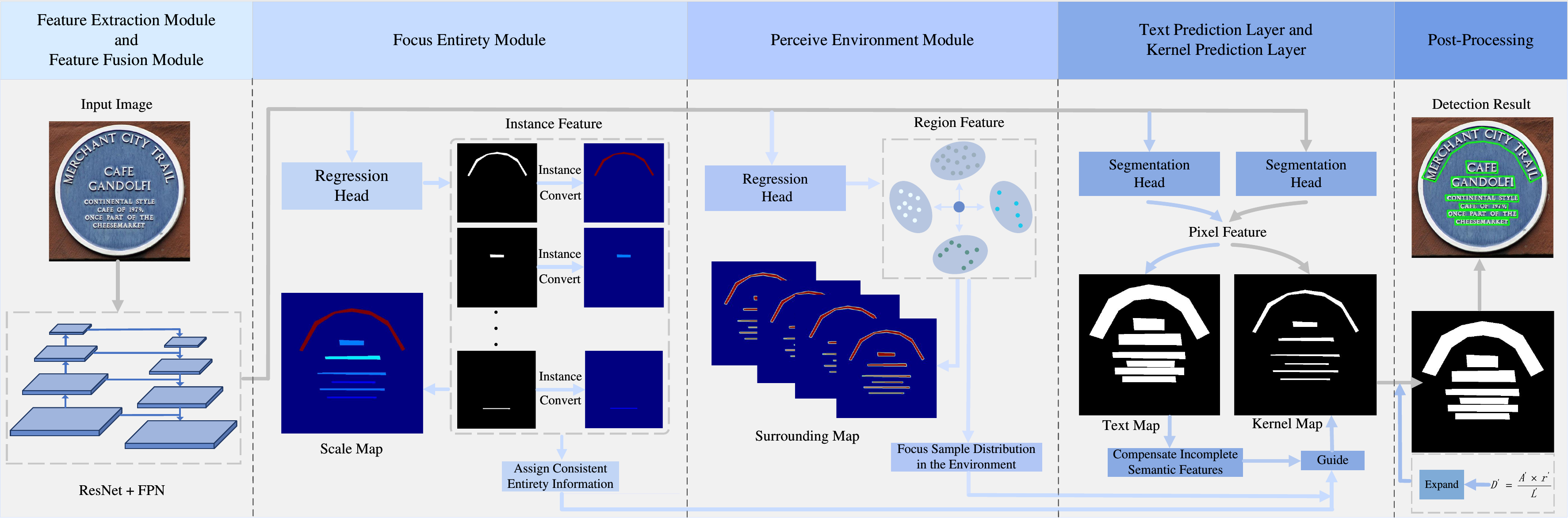}
	\caption{
		The overall framework of the proposed FEPE. During the inference stage, only the feature extraction module, feature fusion module, kernel prediction layer, and post-processing are retained, and the others can be removed. $D^\prime$, $r^\prime$, $A$, and $L^\prime$ represent the expanding distance, expand factor, area, and perimeter of the kernel.}\label{img_overall}
\end{figure*}
\subsection{Segmentation-based methods}

The critical goal of segmentation-based approaches is to predict whether a pixel is text. PSENet \cite{pse} represented text as different scale kernels and reconstructed text instances according to a progressive expansion algorithm. Lyu \emph{et al.} \cite{Lyu_2018_CVPR} proposed an approach that generates candidate boxes by grouping corner points and assess them using region segmentation. Then, the candidate boxes were assessed by region segmentation and suppressed by NMS. TextField \cite{textfield} predicted the direction field while segmenting the text. It utilized the direction field to separate geographically close texts. LeafText \cite{leaftext} treated text instance as leaf and utilized main, lateral, and thin veins to form text. The above approaches perform well for dealing with irregular-shaped texts but still lack efficiency. DBNet \cite{db} segmented the score and regressed the threshold to surprise the result of DB. Benefiting from that DB module can be removed during inference and adopt a lightweight backbone, it achieved excellent performance while maintaining a high inference speed.
On top of that, DBNet++ \cite{db++} introduced an attention mechanism that improves detection accuracy with minimal effect on speed. CM-Net \cite{cm} proposed a novel text kernel representation named concentric mask and learned some auxiliary features to assist in detecting text. PAN \cite{pan} adopted a lightweight backbone and utilized FFM and FPEM to strengthen features. It predicted similar vectors and proposed learnable post-processing for text restructuring.

\section{Method}
\label{method}
 The overall pipeline of the proposed approach is introduced and illustrated first in this section. Then, we describe and visualize the label generation procedure. Furthermore, we present the focus entirety module (FEM) and perceive environment module (PEM) in detail.  Finally, the multi-task constraints loss and training detail are described.

\subsection{Overall Structure}

The overall structure of FEPE is shown in Fig. \ref{img_overall}, which consists of the feature extracting module, feature fusion module, kernel prediction layer, text prediction layer, focus entirety module, and perceive environment module. During the training stage, a multi-level feature map is obtained through the feature extracting and feature fusion modules. Then, the kernel prediction layer, text prediction layer, focus entirety module and perceive environment module output their prediction results. Only the kernel prediction layer is activated during the inference stage, while the text prediction layer, focus entirety module, and perceive environment module are deactivated. The FEM extracts instance-level features and adopts a top-down scheme to model texts which reduces the influence of noises.
Specifically, it assigns consistent information to pixels of the same instance to encourage the clustering of these pixels. In addition, it emphasizes the scale information, enabling the model to distinguish varying scale texts effectively. The PEM defines the kernels as a positive sample and helps the model distinguish the different features between kernels and texts. It defines difference values between edge, internal, and external pixels, thereby helping the model comprehend the distance between contour and pixels. It strengthens the model's comprehension of kernel and text. Benefiting from the superiority of FEM and PEM, when the predictions deviate from the ground truth, these errors tend to be corrected, significantly improving detection accuracy.

The details of the above modules are described as follows. ResNet \cite{resnet} with deformable convolution \cite{deformable1}, \cite{deformable2} is selected as the feature extraction module. Multi-level feature maps are generated through it. The size of feature maps are $\frac{1}{4}$, $\frac{1}{8}$, $\frac{1}{16}$, $\frac{1}{32}$ of the input image, respectively. We adopt the feature pyramid network (FPN) \cite{fpn} to merge multi-level feature maps and obtain the fused feature map $\rm F_f$. This feature map contains both lower-level semantic features and high-level global features. We utilized two segmentation heads with the same architecture for the text kernel prediction layer and text prediction layer, which are shown as follows:
 \begin{equation}
 \rm	H_1= {ReLU_{BN}}( Conv_{3\times3, 64} (F_f)),
 \end{equation}
\begin{equation}
 \rm	H_2= ReLU_{BN} ( ConvT_{3\times3, 64} (H_1)),
\end{equation}
\begin{equation}
 \rm	H_3=  Sigmoid( ConvT_{3\times3, 1} (H_2)),
\end{equation}
where $\rm H_3$, $\rm Conv$, and $\rm ConvT$ represent the output results, convolution operation, and transposed convolution operation, respectively. $\rm {ReLU_{BN}}$ represent the ReLU activation function and batch normalization layer \cite{bn}. The dimension of $\rm H_3$ is $H \times W \times 1$. $H$ and $W$ represent the height and width of the input image.

\begin{figure}[t]
	\centering
	\includegraphics[width=1.0\linewidth,scale=1.0]{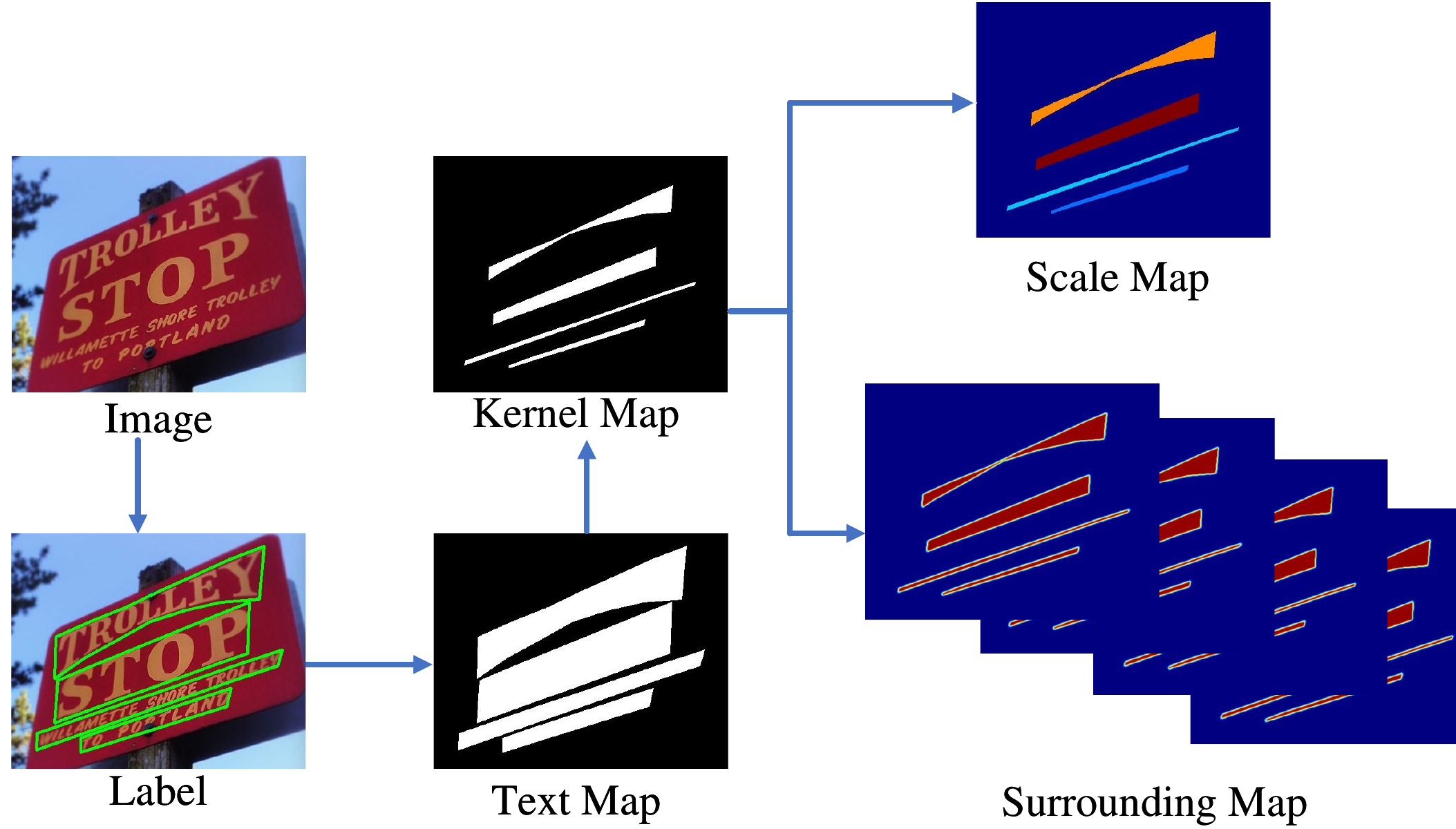}
	\caption{
		The generation process of the text map, kernel map, scale map, and surrounding map used in the experiments.}\label{img_label}
\end{figure}

The label generation processes of the kernel map, scale map, and surrounding map are shown in Fig. \ref{img_label}. The text kernel map is shrunk from the text map, and the shrinkage is calculated based on the area and perimeter of the instance. It can be described in detail as Algorithm \ref{algorithm1}.

%Both the focus entirety module and perceive environment module utilize two similar regression heads, described as follows:
%\begin{equation}
% \rm	W_1= ReLU_{BN} (Conv_{3\times3, 64} (F_f)),
%\end{equation}
%\begin{equation}
% \rm	W_2= ReLU_{BN}( ConvT_{3\times3, 64} (W_1)),
%\end{equation}
%\begin{equation}
% \rm	W_3=  ReLU( ConvT_{3\times3, k} (W_2)),
%\end{equation}
%where $\rm W_3$ is a tensor with size of $H \times W \times 1$ for FEM and $H \times W \times 4$ for PEM.
\begin{algorithm}[t]
	\SetAlgoLined
	\KwData{text map $M_t$,  minimum area threshold $A_{min}$, shrinking ratio $\delta$, area of instance $S$, perimeter of instance $L$, width $W$ and  height $H$}
	\KwResult{ text kernel map $M_k$}
	initializing $M_k\in\mathbb{R}^{W,H}$\;
	\For{${\rm i}$th $\rm instance$ in $M_t$}{
		${\rm offset}_i$ $\leftarrow$ $\frac{S_i}{L_i}( 1-\delta ^2)$\;
		$\rm text$~${\rm kernel}_i$ $\leftarrow$ shrinking contour inward by ${\rm offset}_i$\;
		\If{\rm{area of $\rm text$~${\rm kernel}_i$} $>$ $A_{min}$}{
			drawing $\rm text$~${\rm kernel}_i$ on $M_k$\;
		}
	}
	
	\caption{Text Kernel Label Generation}
	\label{algorithm1}
\end{algorithm}

\subsection{Focus Entirety Module}
\begin{figure}
	\centering
	\label{fee}
	\subfigure[Focus Entirety Module]{
		\begin{minipage}[t]{0.95\linewidth}
			
			\includegraphics[width=1\linewidth]{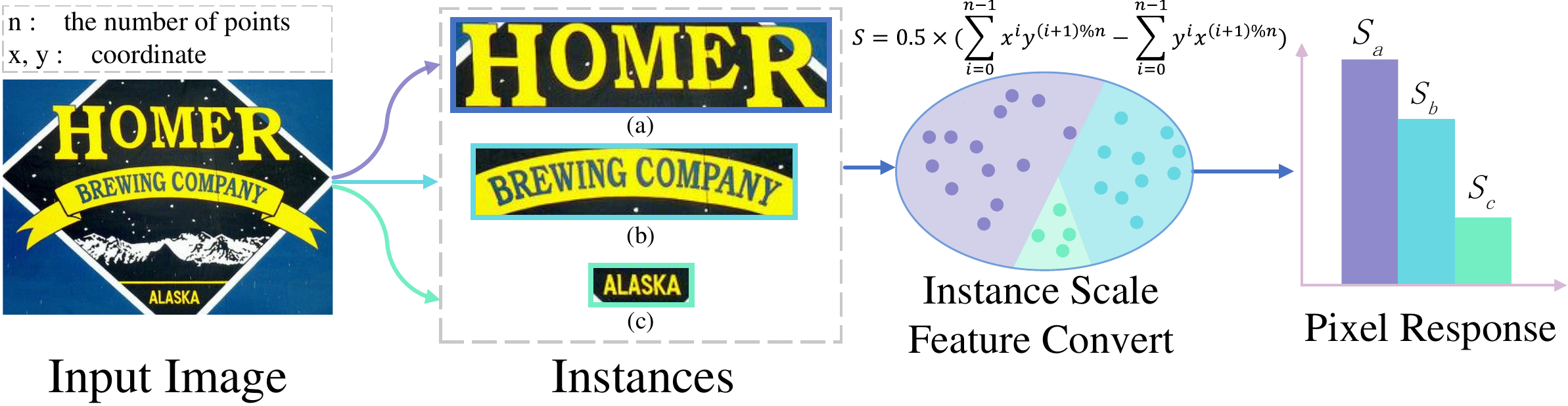}
			
			%\caption{fig1}
		\end{minipage}%
		
	}
	\hspace{-3mm} 
	\label{pe}
	\subfigure[Perceive Environment Module]{
		\begin{minipage}[t]{0.95\linewidth}
			
			\includegraphics[width=1\linewidth]{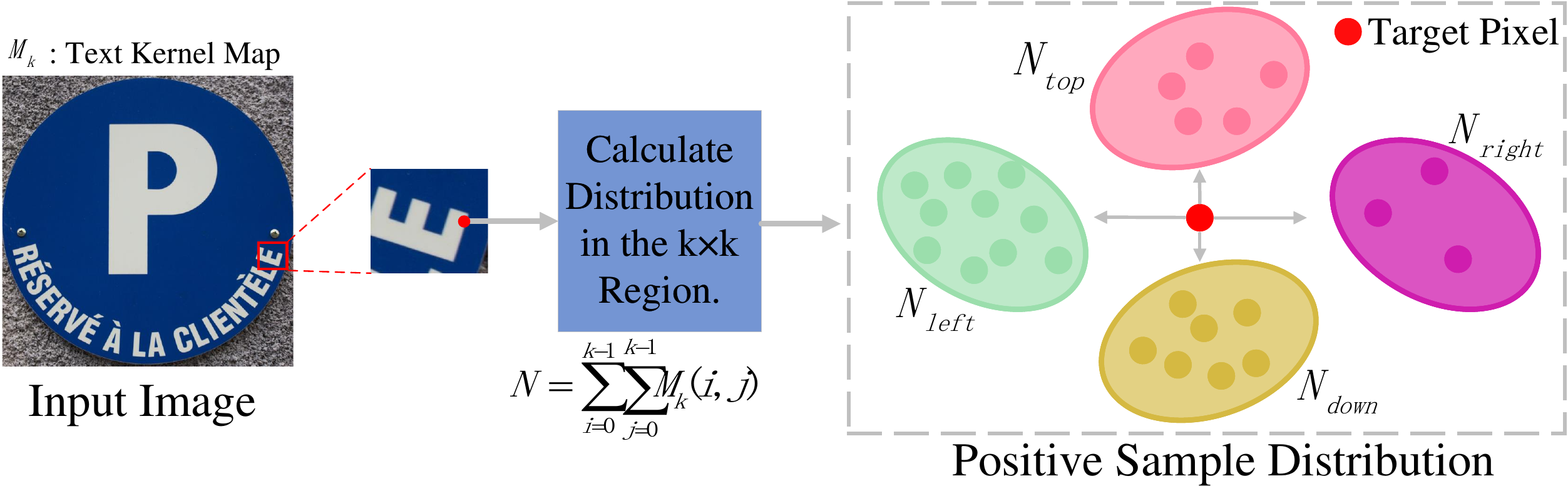}
			%\caption{fig1}
			
		\end{minipage}%
		
	}
	
	\subfigure[An example of FEM]{
		\begin{minipage}[t]{0.50\linewidth}
			
			\includegraphics[width=1\linewidth]{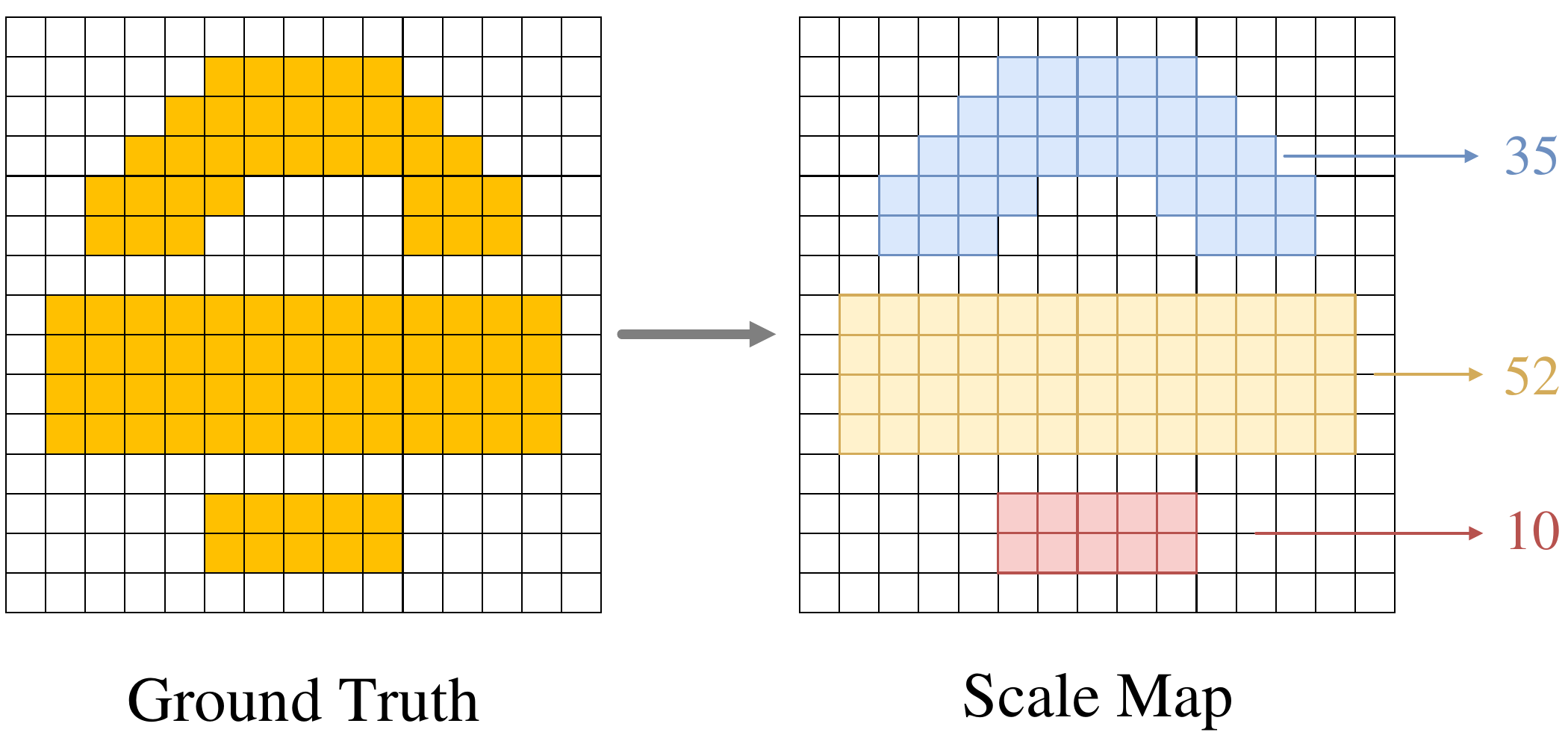}
			%\caption{fig1}
			
		\end{minipage}%
		
	}
	\subfigure[An example of PEM]{
		\begin{minipage}[t]{0.46\linewidth}
			
			\includegraphics[width=1\linewidth]{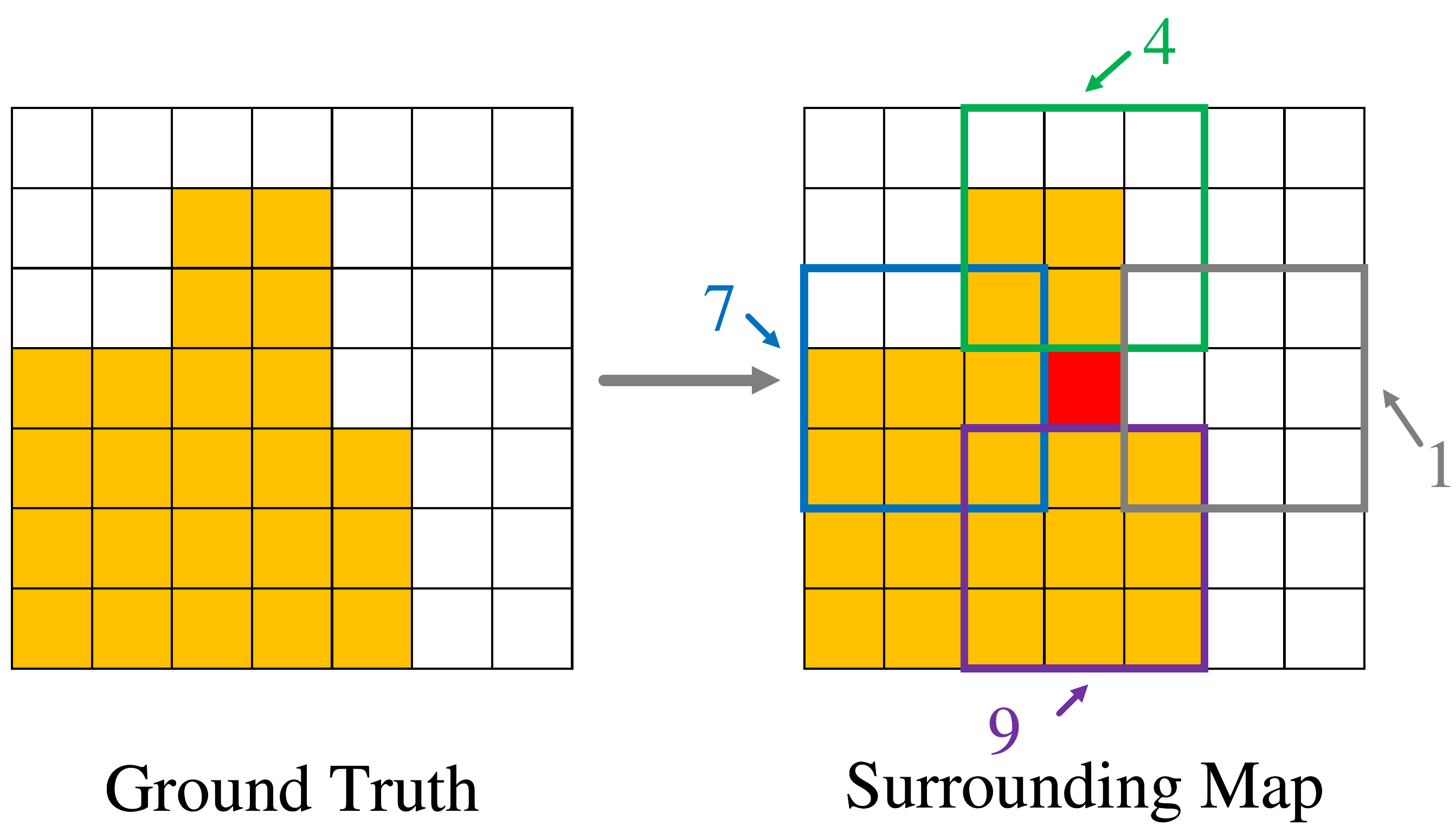}
			%\caption{fig1}
			
		\end{minipage}%
		
	}
	\caption{ The visualization of FEM and PEM. (a) FEM focuses on the scale of instance, the activation value of pixels belonging to large-scale is high. (b) PEM perceives the positive distribution of surroundings. The larger the positive sample the larger the label value. (c) The kernel regions are labeled in orange in the left image. Different instances are marked with a distinct color, and the value is the area of the corresponding instance. (d) The kernel regions are labeled in orange in the left image. The target pixel is marked with red. Its four surrounding map value is generated by the positive pixel number of the purple, green, grey, and blue region.}
	\label{fepe}
\end{figure}

\begin{figure}[t]
	\centering
	\includegraphics[width=1.0\linewidth,scale=1.0]{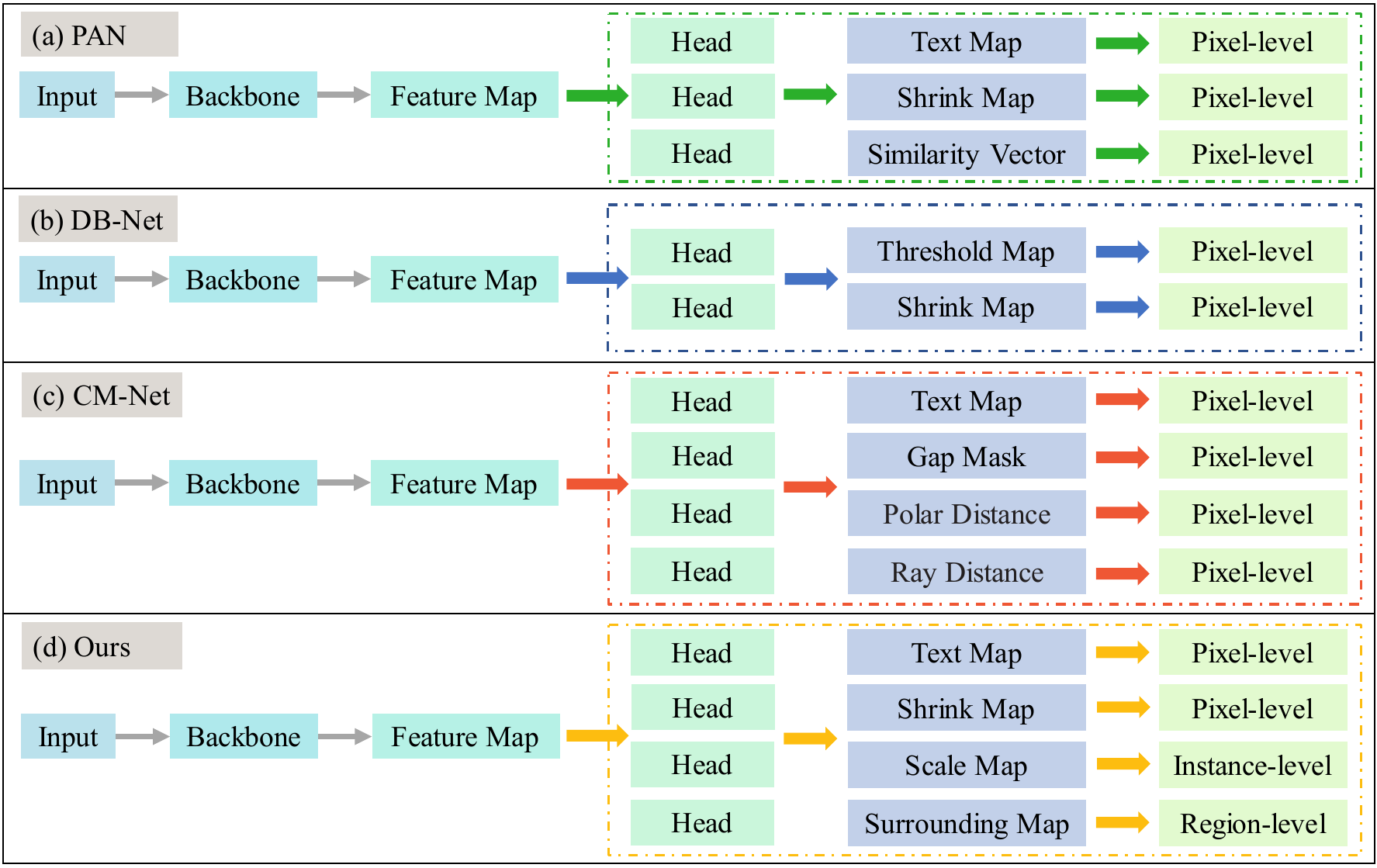}
	\caption{
		The comparison with the overall pipeline of other advanced methods. It provides a comprehensive comparison that describes the hierarchy of features learned by each method.}\label{img_level}
\end{figure}

Segmentation-based approaches are a case of the bottom-up method, which focuses heavily on local information and is susceptible to noise interference. According to this, a focus entirety module (FEM) is proposed to extract instance-level features and model texts in a top-down scheme that reduces the influence of noises.
It assigns consistent entirety information to pixels within the same instance to improve their cohesion and emphasizes the scale of instances, enabling the model to distinguish varying scale texts effectively. The structure of FEM is as follows:
\begin{equation}
	\rm	W_1= ReLU_{BN} (Conv_{3\times3, 64} (F_f)),
\end{equation}
\begin{equation}
	\rm	W_2= ReLU_{BN}( ConvT_{3\times3, 64} (W_1)),
\end{equation}
\begin{equation}
	\rm	W_3=  ReLU( ConvT_{3\times3, k} (W_2)),
\end{equation}
where $\rm W_3$ is a tensor with size of $H \times W \times 1$. 
%It encourages the proposed model that not only predicts the pixel-level information but also predicts the scale map to focus on instance-level information. 
%Notably, the FEM build the instance features and forces pixels on moving toward the instance to which they belong. 
It generates a scale map  $M_{sc}$, which is defined as the area of the corresponding kernel (computed by the Algorithm \ref{algorithm1}). As shown in Fig. \ref{fepe} (a), it focuses on the scale of the instance.
\begin{equation}
	{M^i_{sc}} =
	\begin{cases}
		S^j,  &{\text{if}} \quad  i \in Kernel^j,\\
		{0,}  &{\text{otherwise.}}
	\end{cases}
\end{equation}
where $i$, $S^j$ and $Kernel^j$ represent the $i$th pixel, the area of $j$th kernel and the $j$th kernel. FEM converts the instance scale feature and injects it into the pixel. Pixels belonging to different instances enjoy different response values. The specific generation process of $M_{sc}$ is shown in Algorithm \ref{algorithm}.
As shown in Fig. \ref{img_level}, existing segmentation-based methods reconstruct text instances based on pixel-level knowledge, which lacks instance-level information. Unlike other methods,  FEM infuses the model with information about the scale of the instance to help the pixel determine its attribution. 

\indent

\subsection{Perceive Environment Module}

As exhibited in Fig. \ref{img_level}, existing methods usually focus on information of isolated pixels, such as the probability of being text, kernel \cite{pse}, the similarity vector \cite{pan}, the distance to the text boundary \cite{rmstd}, and the threshold map \cite{db}. However, the local textures of some objects in natural scenes are highly similar to the text. Focusing only on the pixel information is prone to misjudgment.  The visual system tends to rely on surrounding objects to recognize complex objects. Hence, we propose a PEM to provide region-level information about the environment to enhance the model's ability to understand the relative position of pixels in the instance. It effectively improves the accuracy of scoring ambiguous pixels.
The PEM treats the kernel pixels as positive samples to generate surrounding map $M_{sr}$, representing the number of positive samples in the $k\times k$ region in four directions (shown in Fig. \ref{fepe} (b)). It helps the model differentiate text and kernel features, improving the incomplete kernel semantics. The specific generation process of $M_{sr}$ is shown in the Algorithm. \ref{algorithm}.
The structure of PEM is as follows:
\begin{equation}
	\rm	W_1= ReLU_{BN} (Conv_{3\times3, 64} (F_f)),
\end{equation}
\begin{equation}
	\rm	W_2= ReLU_{BN}( ConvT_{3\times3, 64} (W_1)),
\end{equation}
\begin{equation}
	\rm	W_3=  ReLU( ConvT_{3\times3, k} (W_2)),
\end{equation}
where $\rm W_3$ is a tensor with size of $H \times W \times 4$.
When approaching the kernel boundary in a particular direction, the value of the surrounding map corresponding to that direction gradually decreases, which assists in determining the relative position of pixels within the text instance.

 \begin{algorithm}[t]
	\SetAlgoLined
	\KwData{text kernel map $M_k$, area of text kernel instance $S$, environmental perception range $\mu$, width $W$ and height $H$, minimum area threshold $A_{min}$}
	\KwResult{surrounding map $M_{sr}$, scale map $M_{sc}$}
	initializing $M_{sr}\in\mathbb{R}^{W,H,4}$ and $M_{sc}\in\mathbb{R}^{W,H}$\;
	\For{${\rm l}$$th$ $\rm kernel~instance$ in $M_k$}{
		${ \sigma}$ $\leftarrow$ $\rm area~of~ ${\rm l}$th$ $\rm kernel~instance$\;
		\If{$\sigma$ $>$ $A_{min}$}{
			drawing $\rm shrink$-${\rm mask}_k$ on $M_{sc}$ with value ${ \sigma}$\;
		}
	}
		\For{${\rm l}$$th$ ${\rm pixel}_{l}^{i,j}$ in $\rm input~image$}{
			\For{$\rm n$$th$ $\rm M_{sr}$ on ${\rm pixel}_{l}^{i,j}$}{

				$(\theta_{x}^n, \theta_{y}^n)$ $\leftarrow$ $\rm n$$th$ $\rm offset$
				$\rm current~position~(\rho_{x}^{l,n}, \rho_{y}^{l,n})$ $\leftarrow$ $(i,j)+(\theta_{x}^n, \theta_{y}^n)$\;
			
				$\alpha$ $\leftarrow$ clip($\rho_{x}^{l,n}-(\mu+1)/2$, $0$, $W$);
			
				$\beta$ $\leftarrow$ clip($\rho_{x}^{l,n}+(\mu+1)/2$, $0$, $W$);
			
				$\varphi$ $\leftarrow$ clip($\rho_{y}^{l,n}-(\mu+1)/2$, $0$, $H$);
			
				$\omega$ $\leftarrow$ clip($\rho_{y}^{l,n}+(\mu+1)/2$, $0$, $H$);
%				$\rm left,right,top,bottom $ $\leftarrow$ $0,W,0,H$\;
%				\If{$P_{x}^{l,n}-(k+1)/2>0$}{
%					$\rm left$ $\leftarrow$ $P_{ox}-(q+1)/2$\;
%				}
%				\If{$P_{x}^{l,n}+(k+1)/2<W$}{
%					$\rm right$ $\leftarrow$ $P_{ox}+(q+1)/2$\;
%				}
%				\If{$P_{y}^{l,n}-(k+1)/2>0$}{
%					$\rm top$ $\leftarrow$ $P_{oy}-(q+1)/2$\;
%				}
%				\If{$P_{y}^{l,n}+(k+1)/2<H$}{
%					$\rm bottom$ $\leftarrow$ $P_{oy}+(q+1)/2$\;
%				}

				$M_{sr}^{i,j,n}$ 	$\leftarrow$	 $\sum_{m=\alpha}^{\beta} \sum_{v=\varphi}^{\omega} (M_k(m,v)) $;
%			T	$B_{w}^{i,j,n}$ $\leftarrow$ ${\rm OPS}_{n}$\;
			}
		}

	\caption{Scale Map and Surrounding Map Label Generation}
	\label{algorithm}
\end{algorithm}

%\subsection{Label generation}
%The label generation processes of the kernel map, scale map, and surrounding map are shown in Fig. \ref{img_label}. The text kernel map is shrunk from the text map, and the shrinkage is calculated based on the area and perimeter of the instance. It can be described in detail as Algorithm. \ref{algorithm1}.

\subsection{Optimization Function}
In this paper, the proposed FEPE determines four predictions by kernel prediction layer, text prediction layer, focus entirety module, and perceive environment module (as shown in Fig. \ref{img_overall}). A multi-task loss  $\mathcal{L}$ is designed to optimize the proposed method. It includes four loss functions that text segmentation loss $\mathcal{L}_t$, kernel segmentation loss $\mathcal{L}_k$, surrounding map prediction loss $\mathcal{L}_{su}$, scale map prediction loss $\mathcal{L}_{sc}$, which supervise the corresponding features during training.
\begin{equation}
	\mathcal{L}= \lambda_{1}\mathcal{L}_{k}+\lambda_{2}\mathcal{L}_{t} +\lambda_{3}\mathcal{L}_{su} + \lambda_{4}\mathcal{L}_{sc},
\end{equation}
where $\lambda_{1}$, $\lambda_{2}$, $\lambda_{3}$ and $\lambda_{4}$ are the corresponding coefficients of ${L}_{k}$, ${L}_{t}$, ${L}_{su}$, and ${L}_{sc}$. 

%$M_k$
%$N_{right}$
\subsubsection{Kernel Segemention Loss} 
 For kernel segmentation, binary cross-entropy (BCE) loss is utilized for supervision and is often used in binary classification problems.  To alleviate the imbalance of positive and negative samples, hard negative mining is adopted in BCE loss:

%\begin{equation}\label{2}
%	%	\begin{array}{l}
%		S= 0.5 \times (\sum\limits_{i=0}^{n-1}x_i\times y_{(i+1)\% n}-\sum\limits_{i=0}^{n-1}y_i\times x_{(i+1)\% n}) ,
%		%	\end{array}
%\end{equation}
\begin{equation}\label{2}
	%	\begin{array}{l}
		\mathcal{L}_{k}= \sum\limits_{i\in S}-K^{y}_i \times log(K^{x}_i)-(1-K^{y}_i) \times log(K^{x}_i),
		%	\end{array}
\end{equation}
where $S$, $K^{x}$, and $K^{y}$ are the selected training set, prediction kernel map, and ground truth of kernel. The sample ratio between positive and negative samples is 1:3.
\subsubsection{Text Segemention Loss} 

Dice loss is commonly used for segmentation tasks. We adopt it to supervise the text area which is much smaller than the background. The $\mathcal{L}_{t}$ can be describe as follows:
\begin{equation}\label{2}
	%	\begin{array}{l}
		\mathcal{L}_{t} = 1-\frac{2\times \sum_{}(T^y\times T^x)}{\sum_{}T^y+\sum_{}T^x+\varepsilon},
		%	\end{array}
\end{equation}
where $T^y$, $T^x$ are the prediction and ground truth of the text map. $\varepsilon$ is a minimal value used to avoid the denominator being $0$, which is set to $10^{-6}$.

\subsubsection{Regression Loss}
For the surrounding map and scale map, the ratio loss $\mathcal{L}_{ratio}$ \cite{cm} is used to optimize them. It can be described as follows:
\begin{equation}\label{2}
	%	\begin{array}{l}
		\mathcal{L}_{ratio}(X, Y) = log \frac{max(X, Y)}{min(X, Y)},
		%	\end{array}
\end{equation}
where $X$ and $Y$ are the prediction and ground truth, respectively.
$\mathcal{L}_{su}$ and $\mathcal{L}_{sc}$ are based $\mathcal{L}_{ratio}$ which can be defined as follows:
\begin{equation}\label{2}
	%	\begin{array}{l}
		\mathcal{L}_{su}(X_{su}, Y_{su}) = \mathcal{L}_{ratio}(X_{su}, Y_{su}),
		%	\end{array}
\end{equation}
\begin{equation}\label{2}
	%	\begin{array}{l}
		\mathcal{L}_{sc}(X_{sc}, Y_{sc}) = \mathcal{L}_{ratio}(X_{sc}, Y_{sc}),
		%	\end{array}
\end{equation}
where $X_{su}$ and $Y_{su}$ are the prediction and label of the surrounding map, respectively.  $X_{sc}$ and $Y_{sc}$ represent the prediction and label of the scale map, respectively.

%\begin{table}
%	\center
%	\small
%	{
%		\caption{ \small 
%			The detection results on the MSRA-TD500 dataset when $\alpha$ and $\beta$ are initialized to different values.}
%		\begin{tabular}{c|c|ccc}
%			\hline &k$\times$ k  & Precision & Recall & F-measure  \\
%			\hline 
%			LSFL with	&3$\times$3   &84.8 &79.6 &82.1 \\  
%			LSFL with	&5$\times$5  &86.6 &79.0 &82.7 \\ 
%			LSFL with	&7$\times$7  &87.2 &77.5 &82.1 \\ 
%			\hline
%		\end{tabular}
%		\label{tab_ab}
%	}
%\end{table}

\section{Experiment}
\label{experiment}
%First, we describe the data set used in the experiment and the evaluation metrics employed. Then, we conduct ablation experiments on four public datasets to further demonstrate the effectiveness of the proposed method. Immediately after, we compare with other state-of-the-art methods on different datasets. Finally, we analyze the robustness of the method as well as its shortcomings.

In this section, we introduce the datasets. Then,  the ablation studies are conducted on four public benchmarks to prove the superiority of the proposed method. Next,  FEPE is compared with SOTA methods on different public benchmarks. Finally, we demonstrate the robustness of the method, and its shortcomings are also analyzed.

\begin{table*}[!t]
	\center
	
 {
		\caption{ Ablation study on the effect of FEM and PEM on detection performance on the ICDAR2015 and MSRA-TD500. ``Text'' represents the text prediction layer. }
		\centering
		{ 
			\begin{tabular}{c|ccc|ccc|ccc}
				
				\hline
				\multirow{2}{*}{Backbone}  &\multirow{2}{*}{Text } 
				&\multirow{2}{*}{FEM} &\multirow{2}{*}{PEM }
				& \multicolumn{3}{c}{MSRA-TD500}	        &\multicolumn{3}{c}{ICDAR2015}  \\    
				% & Min  &Average  & Max\\	
				\cline{5-10} 
				&  & & &Precision &Recall &F-measure  &Precision &Recall &F-measure \\ \hline 
				
				\multirow{7}{*}{ResNet18} 
				&$\times$ &$\times$ &$\times$ &79.4 &76.8 &78.1   &87.7 &75.2 &81.0  \\ 
				%			Sigmoid($\alpha \times P_t- \alpha \times P_k$) &res18   &86.6 &84.5 &85.6 &12.3 &87.2 &80.0 &83.4  &0\\ 
				&\checkmark &$\times$ &$\times$ & 80.1 &77.0 &78.5  &89.2 &74.6 &81.3 \\ 
				&$\times$ &\checkmark &$\times$  &84.1 &80.8 &82.4   &88.4 &78.3 &83.0 \\ 
				&$\times$  &$\times$ &\checkmark &82.9 &79.7&81.3 &88.2 &78.5 &83.1 	\\ 
				&\checkmark &\checkmark &$\times$  &87.0 &79.4 &83.0   &88.5 &77.1 &82.4 \\ 
				&\checkmark  &$\times$ &\checkmark &86.2 &79.2&82.5 &88.0 &78.5 &83.0 	\\ 
				
				&\checkmark  &\checkmark &\checkmark &87.7 &80.6&84.0  &87.3 &79.4 &83.2  	\\ \hline
				\multirow{7}{*}{ResNet50}   &$\times$ &$\times$ &$\times$ &85.8 &79.9 &82.7  &87.5 &79.5 &83.3  \\ 
				%			Sigmoid($\alpha \times P_t- \alpha \times P_k$) &res18   &86.6 &84.5 &85.6 &12.3 &87.2 &80.0 &83.4  &0\\ 
				&\checkmark &$\times$ &$\times$ &86.2 &82.3 &84.2  &89.3 &78.3 &83.4  \\ 
				&$\times$ &\checkmark &$\times$  &86.9 &83.3 &85.1   &88.2 &80.5 &84.1 \\ 
				&$\times$  &$\times$ &\checkmark &89.0 &82.0&85.3 &88.2 &80.4 &84.1 	\\ 
				&\checkmark &\checkmark &$\times$ &90.2 &80.1 &85.0  &87.1&81.1 &84.0  \\ 
				
				&\checkmark &$\times$ &\checkmark  &89.2 &82.0 &85.4  &88.0 &80.0 &83.8  	\\ 
				&\checkmark  &\checkmark &\checkmark&88.1 &83.5 &85.8  &88.5 &80.4 &84.2	\\
				%			\checkmark &\checkmark &BITM$\_$V2 &\checkmark &87.7 &79.6 &83.4 &62 &89.7 &76.7 &82.7 &48\\
				\hline
			\end{tabular}
			
		}
		\label{tab_abs1}	
		
	}
\end{table*}
\begin{table*}[!t]
	\center
	
 {
		\caption{  Ablation study on the effect of FEM and PEM on detection performance on the Total-Text and CTW1500. ``Text'' represents the text prediction layer.}
		\centering
		{ 
			\begin{tabular}{c|ccc|ccc|ccc}
				
				\hline
				\multirow{2}{*}{Backbone}  &\multirow{2}{*}{Text } 
				&\multirow{2}{*}{FEM} &\multirow{2}{*}{PEM }
				& \multicolumn{3}{c}{Total-Text}	        &\multicolumn{3}{c}{CTW1500}  \\    
				% & Min  &Average  & Max\\	
				\cline{5-10} 
				&  & & &Precision &Recall &F-measure  &Precision &Recall &F-measure \\ \hline 
				
				\multirow{7}{*}{ResNet18} 
				&$\times$ &$\times$ &$\times$ &86.0 &76.4 &80.9   &81.9 &79.4 &80.6  \\ 
				%			Sigmoid($\alpha \times P_t- \alpha \times P_k$) &res18   &86.6 &84.5 &85.6 &12.3 &87.2 &80.0 &83.4  &0\\ 
				&\checkmark &$\times$ &$\times$ & 87.2 &78.3 &82.5 &82.6 &80.8 &81.7 \\ 
				&$\times$ &\checkmark &$\times$  &85.2 &80.4 &82.7 &83.4 &80.6 &82.0    \\ 
				&$\times$  &$\times$ &\checkmark &87.1 &78.9 &82.8 &83.8 &80.9 &82.3  	\\ 
				&\checkmark &\checkmark &$\times$  &87.6 &78.6 &82.9   &84.3 &81.4 &82.9 \\ 
				&\checkmark  &$\times$ &\checkmark &87.1 &79.1&82.9 &83.7 &81.4 &82.6	\\ 
				&\checkmark  &\checkmark &\checkmark &89.4 &78.8 &83.7  &85.1 &81.6 &83.3  	\\ \hline
			\end{tabular}
			
		}
		\label{tab_abs2}	
		
	}
\end{table*}
\subsection{Datasets}
\textbf{CTW1500}~\cite{yuliang2017detecting}  is a text dataset includes long curved text. It contains 1,000 training images and 500 testing images. 
Each text instance is labeled with 14 points.

\textbf{ICDAR2015}~\cite{karatzas2015icdar} contains many samples from supermarkets, and many of these examples have low-resolution problems. Each instance is labeled by a quadrilateral consisting of four points. It contains 1500 images.

\textbf{Total-Text}~\cite{ch2017total} contains not only a large amount of horizontal text and multi-directional text but also a large amount of irregularly shaped text. The training and test sets have 1255 and 300 images, respectively.

\textbf{SynthText}~\cite{gupta2016synthetic} is a synthetic text dataset that includes 800,000 images for pre-training. It is generally used to pre-train the model to improve its performance.

\textbf{ICDAR2017 MLT} is a multilingual text dataset. It consists of 7,200 training images, 1,800 validation images, and 9,000 testing images in nine languages.

\textbf{MSRA-TD500}~\cite{yao2012detecting} is a  Chinese-English bilingual scene text dataset with line-level annotations. We follow previous papers to utilize HUST-TR400 \cite{yao2014unified} for training.

%\subsection{Evaluation Metrics}
\subsection{Implementation Details}   
ResNet with deformable convolution and Feature Pyramid Network (FPN) are selected as the backbone. We choose two pre-training strategies: (1) Pretraining on ICDAR2017MLT for 400 epochs. (2) Pretraining on SynthText for four epochs. Afterward, the model is fine-tuned for 1,200 epochs. During the training phase, the batch size and initial learning rate are set to 16 and 0.007, respectively. The stochastic gradient descent (SGD) is used to train the model, while the weight decay and momentum are set to 0.0001 and 0.9, respectively. We use the ``poly" strategy to adjust the learning rate, where the current learning rate is equal to the initial learning rate multiplied by $(1-\frac{iter}{max\_iter})^{power}$, and the power is set to 0.9. Slight random rotation, random cropping, and random flipping are used for data augmentation. All input images are resized to 640$\times$640 during training. We evaluated the detection results following the metrics used in DBNet. During the inference stage, the prediction of the kernel map is binarized, and each kernel instance is obtained through contour extraction. Then, each text kernel expands a specific distance $D^\prime = \frac{A^\prime \times r^\prime}{L^\prime}$ to generate the text instance.  $r^\prime$, $A^\prime$, and $L^\prime$ represent the expand ratio, area, and perimeter of the kernel. The coefficients of the loss $\lambda_{1}$, $\lambda_{2}$, $\lambda_{3}$ and $\lambda_{4}$ are set to 6, 3, 1, and 0.5, respectively.
\subsection{Ablation Study}  
The ablation study is conducted on four public benchmarks to show the effectiveness of the proposed PEM and FEM.
All models are trained without pre-training.

\begin{table*}[!t]
	\center
 {
		\caption{  Ablation study on the impact of $k$ on detection performance on the ICDAR2015 and MSRA-TD500.   $k$  represents the area perceived by PEM in the range of $k \times k$.}
		\centering
		{ 
			\begin{tabular}{c|c|cccc|cccc}
				
				\hline
				\multirow{2}{*}{} &\multirow{2}{*}{Kernel} 
				& \multicolumn{4}{c}{MSRA-TD500}	        &\multicolumn{4}{c}{ICDAR2015}  \\    
				% & Min  &Average  & Max\\	
				\cline{3-10} 
				& &Precision &Recall &F-measure &FPS &Precision &Recall &F-measure &FPS\\ \hline 
				
				\multirow{3}{*}{FEPE with} 
				&3$\times$3  &87.6 &77.8 &82.4 &62  &88.0 &78.5 &83.0 &48 \\ 
				&5$\times$5  &87.0 &79.4 &83.0 &62 &88.9 &77.8 &83.0 &48 \\ 
				&7$\times$7 &84.7 &77.5 &81.6 &62 &89.0 &77.4 &82.8 &48  	\\ \hline
				
			\end{tabular}
			
		}
		\label{tab_k}	
		
	}
\end{table*}

\begin{table*}[!t]
	\center
	
{
		\caption{ Ablation study on the impact of $k$ on detection performance on the Total-Text and CTW1500.   $k$  represents the area perceived by PEM in the range of $k \times k$.}
		\centering
		{ 
			\begin{tabular}{c|c|cccc|cccc}
				
				\hline
				\multirow{2}{*}{} &\multirow{2}{*}{Kernel} 
				& \multicolumn{4}{c}{Total-Text}	        &\multicolumn{4}{c}{CTW1500}  \\    
				% & Min  &Average  & Max\\	
				\cline{3-10} 
				& &Precision &Recall &F-measure &FPS &Precision &Recall &F-measure &FPS\\ \hline 
				
				\multirow{3}{*}{FEPE with} 
				&3$\times$3  &88.0 &77.6 &82.4 &50  &84.5 &80.5 &82.5 &55 \\ 
				&5$\times$5  &87.1 &79.1 &82.9 &50 &83.7 &81.4 &82.6 &55 \\ 
				&7$\times$7 &87.5 &77.3 &82.1 &50 &82.9 &81.5 &82.2 &55  	\\ \hline
				
			\end{tabular}
			
		}
		\label{tab_curve_k}	
		
	}
\end{table*}

\begin{table}
	
	\center
{
		\caption{ The qualitative analysis of whether the model enhancement is due to the extra supervision, where 'Kernel', 'Scale', and 'Surrounding' represent the kernel map, scale map, and surrounding map.}
		\begin{tabular}{cccccc}
			
			\hline Baseline &PEM &FEM & P &R &F \\
			\hline Kernel &- &- &79.4 &76.8 &78.1 \\
			Kernel	&Kernel &- &81.4 & 79.2 &80.3\\
			Kernel	&Scale &- &87.0 & 79.4 &83.0\\
			Kernel	&- &Kernel &80.4 & 74.6 &77.5\\
			Kernel	&- &Surrounding &86.2 & 79.2 &82.5\\
			%			\hline \multirow{3}{*}{MSRA-TD500} &None &87.7 &80.6 &84.0 \\
			%			&SynthText & 89.4 &82.8 &86.0\\
			%			&ICDAR2017 & 93.6 & 85.4 &89.3\\
			\hline
		\end{tabular}
		\label{tab_extra}
	}
	
\end{table}

\subsubsection{Effectiveness of the FEM}
\begin{figure}[t]
	\centering
	\includegraphics[width=0.95\linewidth,scale=1.0]{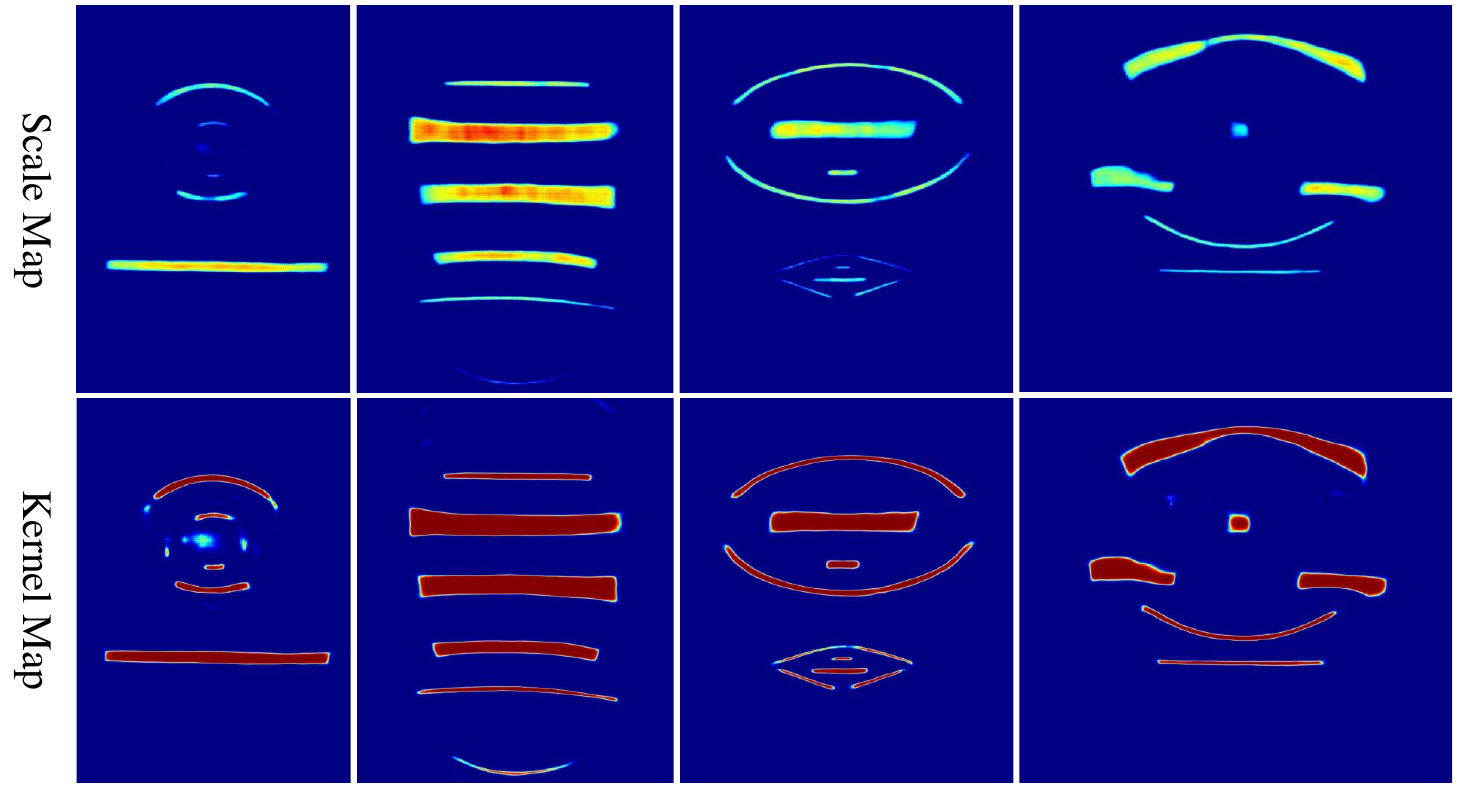}
	\caption{
	The visualization of the prediction of the scale map and the corresponding kernel map. For the scale map, a redder color means a higher value.}\label{img_area}
\end{figure}

The proposed FEM assigns consistent entirety information to pixels within the same instance to improve their cohesion and emphasizes the scale of instances to which pixels belong, enabling the model to distinguish varying scale texts effectively. Extensive experiments have proved the superiority of the proposed FEM. As seen in Table \ref{tab_abs1}, the proposed FEM brings a 4.3$\%$ and 2.0$\%$ performance improvement on MSRA-TD500 and ICDAR2015 when ResNet18 is adopted as the backbone. For ResNet50, the improvement of the method is 2.4$\%$ and 0.8$\%$, respectively. In addition, when adopting ResNet18 as the backbone, FEM yields about 1.8$\%$ and 1.4$\%$ improvement on Total-Text and CTW-1500, respectively.  The above experiment results demonstrate the superiority of the proposed FEM. As shown in Fig. \ref{img_area}, we display the prediction of the scale map that the redder means a higher value. As we can see, the larger instance obtains the higher value, which demonstrates the proposed FEM modeling the instance feature successfully. Moreover, the model's predictions for the scale map and kernel map are relatively consistent.

\begin{table}
	
	\center
{
		\caption{ The detection performance of FEPE with different pre-training conditions on four public benchmarks.}
		\begin{tabular}{ccccc}
			
			\hline Datasets &Ext. & P &R &F \\
			\hline \multirow{3}{*}{TotalText} &None &89.4 &78.7 &83.7 \\
			&SynthText & 90.8 & 79.5 &84.8\\
			&ICDAR2017 & 89.2 & 79.2 &83.9\\
			\hline \multirow{3}{*}{MSRA-TD500} &None &87.7 &80.6 &84.0 \\
			&SynthText & 89.4 &82.8 &86.0\\
			&ICDAR2017 & 93.6 & 85.4 &89.3\\
			\hline \multirow{3}{*}{CTW1500} &None &85.1 &81.6 &83.3 \\
			&SynthText & 88.0 & 83.0 &85.5\\
			&ICDAR2017 & 89.0 & 82.2 &85.5\\
			\hline \multirow{3}{*}{ICDAR2015} &None &88.0 &78.5 &83.0 \\
			&SynthText & 87.3 & 79.4 &83.2\\
			&ICDAR2017 & 89.9 & 79.7 &83.5\\
			\hline
		\end{tabular}
		\label{tab_data}
	}
	
\end{table}

\subsubsection{Influence of the PEM}
\begin{figure}[t]
	\centering
	\includegraphics[width=0.95\linewidth,scale=1.0]{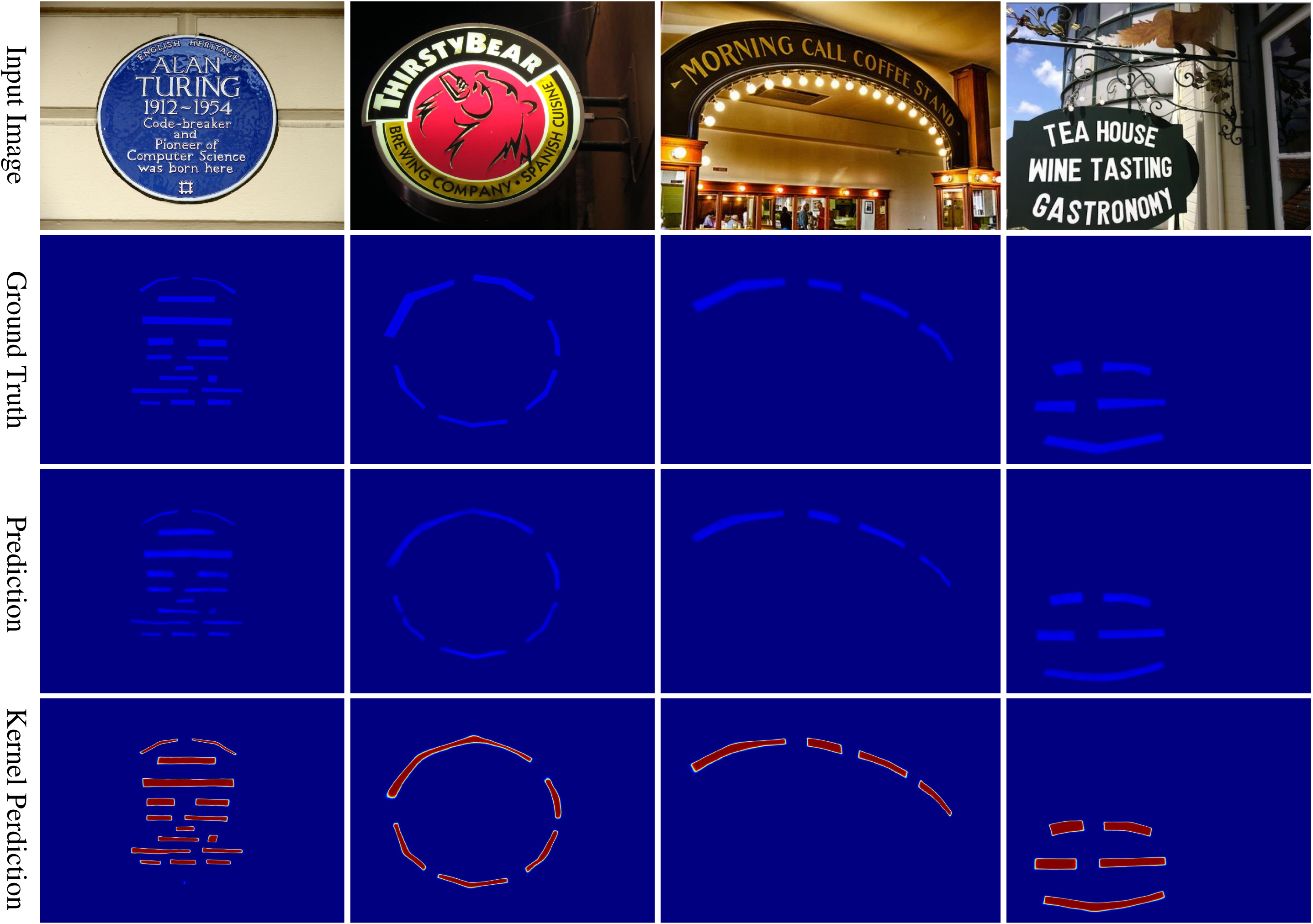}
	\caption{
		Visualization of left surrounding maps predictions and ground truths, as well as kernel maps predictions.}\label{img_pos}
\end{figure}

As mentioned above, PEM extracts region-level features and encourages the model to focus on the distribution of positive samples in the vicinity of a pixel, which perceives environment information. It treats the kernel pixels as positive samples and helps the model differentiate text and kernel features, improving the incomplete kernel semantics. A series of experiments are conducted on MSRA-TD500, Total-Text, CTW1500, and ICDAR2015 datasets to validate the superiority of the proposed PEM. As shown in Table \ref{tab_abs1}, the proposed PEM improved F-measure by 3.2$\%$ and 2.1$\%$ on MSRA-TD500 and ICDAR2015 when ResNet18 is used as the backbone. Moreover, the method brings 2.6$\%$ and 0.8$\%$ performance improvements when using ResNet50 to extract features. For Total-Text and CTW1500, when adopting ResNet18 as the backbone, the proposed PEM achieves 1.9$\%$ and 1.7$\%$ performance gains, respectively, as shown in Table \ref{tab_abs2}. The above experiments demonstrate that PEM can help the proposed model to improve detection performance effectively. As shown in Fig. \ref{img_pos}, we show the prediction of the left surrounding map, corresponding ground truth, which actually models regional features of texts. In addition, the prediction of the left surrounding map and the kernel map are consistent.
\subsubsection{Influence of the choice of $k$}
Table \ref{tab_k} and Table \ref{tab_curve_k} show the results under different $k \times k$ regions to validate the impact of $k$ on detection performance on ICDAR2015, MSRA-TD500, Total-Text and CTW1500, respectively. When $k$ is set to 5, the model achieves optimal performance, and we set it up like this in subsequent experiments. When the choice of $k$ is too large, it will occupy too much weight of the model, and when it is too small, it will affect the model's ability to perceive the environment.
\subsubsection{The qualitative analysis of the extra supervision.}
To qualitatively analyze whether the model enhancement is due to the extra supervision, we perform corresponding experiments on MSRA-TD500. We change the label of FEM and PEM to kernel map. As shown in Table \ref{tab_extra}, the extra supervision for the model is not always helpful for the model. Compared with using kernel maps, the scale map and surrounding map are better, which shows that the effectiveness of FEM and PEM is not because of extra supervision.

\indent
\begin{figure}[t]
	\centering

	\includegraphics[width=0.9\linewidth,height=5.7cm]{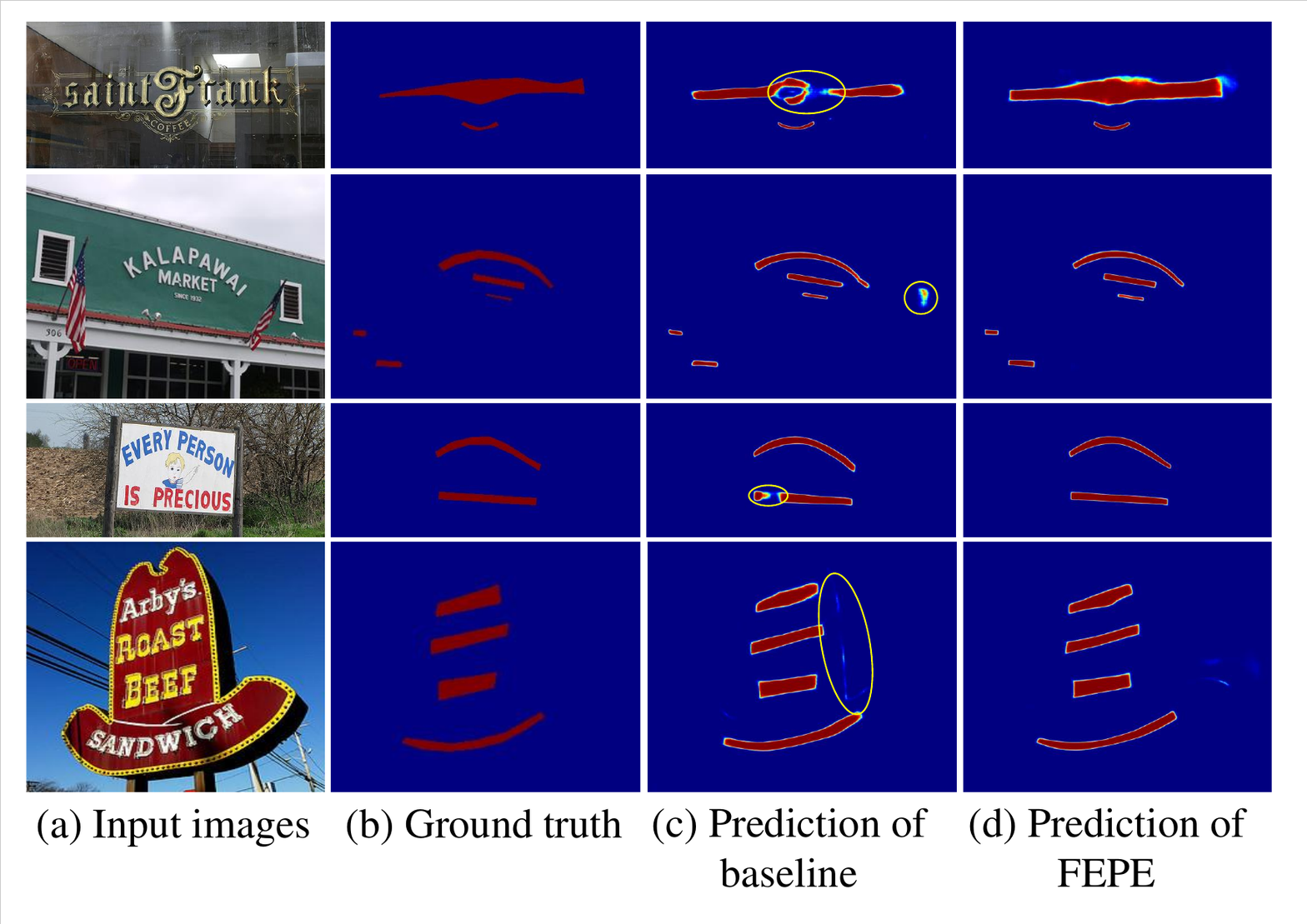}
	
	\caption{ The visual comparison between the ground truth (text kernel maps) and the predicted results. The input images are presented in (a). The ground truth is illustrated in (b), while (c) and (d) show the prediction results of the baseline (only predict text kernel maps and text maps) and the proposed FEPE, respectively. The baseline method faces difficulties correctly identifying long text regions with large gaps, which are misjudged as non-text regions in (c). Moreover, patterns resembling text texture are mistakenly classified as text regions, as shown in (c). 
		%		However, the proposed method shows improved accuracy in correctly detecting text regions and overcoming the aforementioned issues.
	}
	\label{first}
\end{figure}

\begin{table*}[ht]
	\center
	\renewcommand\arraystretch{1.0}
	\setlength{\tabcolsep}{2.5mm}

		\caption{ Comparison with existing advanced approaches on the MSRA-TD500 and CTW1500 datasets. ``\textcolor{red}{\textbf{Red}}'', \textcolor{blue}{\textbf{Blue}}'' and  ``\textcolor{green}{\textbf{Green}}'' represent the optimal, sub-optimal and  the third best performance, respectively.}
		
		{ 
			\begin{tabular}{cccccccccccc}
				
				\hline
				
				\multirow{2}{*}{Methods} &\multirow{2}{*}{Venue} &\multirow{2}{*}{Ext.} &\multirow{2}{*}{Backbone }
				& \multicolumn{4}{c}{MSRA-TD500}	        &\multicolumn{4}{c}{CTW1500}  \\    
				% & Min  &Average  & Max\\	
				\cline{5-12} 
				& & & &P &R &F &FPS &P &R &F &FPS\\ \hline

				%				PSE-1s\cite{pse}  &CVPR’19&MLT &ResNet50&- &- &- &- &84.8 &79.7 &82.2 &3.9\\ 
				PAN\cite{pan} &ICCV’19 &Synth &ResNet18 &84.4 &83.8 &84.1 &30.2 &86.4 &81.2 &83.7 &39.8\\ 
				ContourNet\cite{contournet}  &CVPR'20 &- &ResNet50 &- &- &- &- &84.1 &83.7 &83.9 &4.5\\ 
				DRRG\cite{zhang2020deep} &CVPR’20 &MLT &VGG16 &88.1 &82.3 &85.1 &- & 85.9 &83.0 &84.5 &- \\  	 
				%				DBNet\cite{db} &AAAI’20&Syn&ResNet50 &{91.5} &{79.2} &{84.9} &{32} &{86.9} &{80.2} &{83.4} &{22}\\ 
				%				DBNet\cite{db} &AAAI’20&Syn&ResNet18 &{90.4} &{76.3} &{82.8} &\textcolor{red}{\textbf{62}} &{84.8} &{77.5} &{81.0} &\textcolor{red}{\textbf{55}}\\	
				
				%FCENet\cite{fcenet} &CVPR’21&- &ResNet50 &- &- &- &- &87.6 &83.4 &{85.5} &-\\ 	 
				
				CTNet\cite{ct}&NeurIPS’21 &Synth &ResNet18 &90.0 &82.5  &86.1 &34.8 &{88.3} &79.9 &83.9 &40.8\\ 
				
				FEMP\cite{femp} &TMM'21 &MLT &ResNet50&86.0 &83.4 &84.7 &1.6 &88.5 &82.9 &\textcolor{blue}{\textbf{85.6}} &1.4 \\
				
				PCR\cite{pcr}  &CVPR’21 &MLT &DLA34 &90.8 &83.5 &87.0 &- &87.2 &82.3 &84.7 &-\\ 	 
				%			MOST\cite{he2021most} &Res50  &{90.4} &{82.7} &{86.4} &{51.8} &{-} &{-} &{-} &{-}\\
				TextBPN\cite{textbpn}&ICCV’21 &Synth  &ResNet50   &85.4 &80.7 &83.0 &12.7 &87.8 &{81.5} &84.5 &12.2 \\ 
				TextBPN\cite{textbpn}&ICCV’21 &MLT &ResNet50   &86.6 &84.5 &85.6 &12.3 &86.5 &{83.6} &85.0 &12.2 \\  	
				LPAP\cite{lpap} &TOMM'22 &Synth &ResNet50 & 87.9 &77.7 &82.5 &- &84.6 &80.3 &82.4 &- \\
				
				%FEMP &TMM'21 &MLT &ResNet50&86.0 &83.4 &84.7 &1.6 &88.5 &82.9 &85.6 &1.4 
				
				TextDCT \cite{textdct} &TMM'22 &Synth &ResNet50 &- &- &- &- & 85.3 &85.0 &85.1 &17.2 \\
				ASTD \cite{astd} &TMM'22 &- &ResNet101 &- &- &- &- &87.2 &81.7&84.4 &- \\
				LEMNet \cite{xing2021boundary} &TMM'22 &- &ResNet50 &85.6 &84.8 &85.2 &- & 86.6 &83.8 &85.2 &- \\
				ADNet\cite{ad} &TMM’22 &Synth &ResNet50 &92.0 &83.2 &87.4 &- &88.2 &83.1 &\textcolor{blue}{\textbf{85.6}} &-\\ 
				CMNet\cite{cm} &TIP’22 &- &ResNet18 &89.9 &80.6 &85.0 &41.7 &86.0 &82.2 &84.1 &\textcolor{green}{\textbf{50.3}}\\  
%					Tang \emph{et al.} \cite{tang2022few} &CVPR’22 &Synth &ResNet50  &{91.6} &{84.8} &\textcolor{blue}{\textbf{88.1}} &{-} &{88.1} &{82.4} &{85.2} &{-}\\ 
				PAN++ \cite{pan++} &TPAMI'22 &Synth &ResNet18 &89.6 &86.3 &\textcolor{green}{\textbf{87.9}} &22.6 &87.1 &81.1 &84.0 &36.0\\
				KPN\cite{9732893}&TNNLS’22 &MLT &ResNet50  &{-} &{-} &{-} &{-} &{84.4} &{84.2} &84.3 &{16}\\ 
				ZTD \cite{zoom} &TNNLD'23 &Synth &ResNet18 &91.6 &82.4 &86.8 &\textcolor{blue}{\textbf{59.2}} &88.4 &80.2 &84.1 &\textcolor{red}{\textbf{76.9}}\\
				FS \cite{9870672} &TIP’23 &- &ResNet18  &{90.0} &{80.4} &{84.9} &{35.5} &{84.6} &{77.7} &{81.0} &{35.2}\\ 
				FS \cite{9870672} &TIP’23 &- &ResNet50  &{89.3} &{81.6} &{85.3} &{25.4} &{85.3} &{82.5} &{83.9} &{25.1}\\ 
				DBNet++\cite{db++}&TPAMI’23&Synth &ResNet18 &{87.9} &{82.5} &{85.1} &\textcolor{green}{\textbf{55}} &{84.3} &{81.0} &{82.6} &49\\
				DBNet++\cite{db++}&TPAMI’23&Synth &ResNet50  &{91.5} &{83.3} &{87.2} &{29} &{87.9} &{82.8} &{85.3} &{26}\\ 
			
				LeafText \cite{leaftext} &TMM'23 &Synth &ResNet50/18 &92.1 &83.8 &86.1 &-  &87.1 &83.9 &\textcolor{green}{\textbf{85.5}} &-\\ \hline
				%			XXX &2022 TGRS &7.93 &11.82 &238.46 &625.90 &12.94 &20.25 &50.45 &65.24 \\ 	\hline
				\textbf{FEPE} &Ours &Synth &ResNet18 &{89.4} &{82.8} &86.0 &\textcolor{red}{\textbf{62}} 
				&{88.0} &{83.0} &\textcolor{green}{\textbf{85.5}} &\textcolor{blue}{\textbf{55}}\\ 	
				
				\textbf{FEPE} &Ours &MLT &ResNet18 &{93.8} &{85.6} &\textcolor{red}{\textbf{89.5}} &\textcolor{red}{\textbf{62}} &{89.0} &{82.2} &\textcolor{green}{\textbf{85.5}} &\textcolor{blue}{\textbf{55}}\\ 
				%			SSFL &- &Syn &Res50 &$\mathbf{93.3}$ &$\mathbf{86.3}$ &$\mathbf{89.6}$ &$\mathbf{62}$ &{89.2} &{82.5} &{85.7} &$\mathbf{55}$\\ 
				\textbf{FEPE} &Ours &Synth &ResNet50 &{90.5} &{85.4} &\textcolor{blue}{\textbf{88.0}} &32 &{88.8} &83.5 &\textcolor{red}{\textbf{86.0}} &22\\ 	
				%			SSFL &- &MLT &Res50 &$\mathbf{93.3}$ &$\mathbf{86.3}$ &$\mathbf{89.6}$ &$\mathbf{62}$ &{89.2} &{82.5} &{85.7} &$\mathbf{55}$\\ 
				\hline
			\end{tabular}
		}
		\label{td500}
	
\end{table*}

\begin{table}
	\center

	{
		\caption{  Comparison with existing advanced approaches on the Total-Text. ``\textcolor{red}{\textbf{Red}}'' and ``\textcolor{blue}{\textbf{Blue}}'' represent the optimal and sub-optimal performance, respectively.}
		\begin{tabular}{cccccc}

			\hline Methods &Backbone & P & R & F & FPS \\
			\hline 
			
			%EAST\cite{east}  & $83.6$ & $73.5$ & $78.2$ & $13.2$ \\
			%TextBoxes++\cite{liao2018textboxes++} & $87.2$ & $76.7$ & $81.7$ & $11.6$ \\
			PSENet-1s\cite{pse} &ResNet50& 84.0 & 78.0 & 80.9 & 3.9 \\
			TextSnake\cite{textsnake} &VGG16 & 82.7 & 74.5 & 78.4 &- \\
			%	\cite{tang2022few}  & $\mathbf{90.7}$ & $\mathbf{85.7}$ & $\mathbf{88.1}$ & $-$ \\
			Boundary\cite{wang2020all} &ResNet50& 85.2 & 82.2 & 84.3 & - \\
			DRRG\cite{zhang2020deep} &VGG16& 86.5 & {84.9} & 85.7 & - \\
			FCENet\cite{fcenet}&ResNet50 & {89.3} & 82.5 & 85.8 & - \\
			KPN\cite{9732893} &ResNet50& 88.0 & {82.3} & 85.1 & 22.7 \\
			%			DB++\cite{db++} &ResNet50&{88.9} & {83.2} & \textcolor{blue}{\textbf{86.0}} & {28} \\
			PSE+STKM\cite{wan2021self} &ResNet50 &86.3 &78.4 &82.2 &- \\
			DB\cite{db} &ResNet50 & {87.1} & 82.5 & {84.7} &32 \\
			CM-Net\cite{cm} &ResNet18& 88.5 & {81.4} & 84.8 & \textcolor{blue}{\textbf{49.8}} \\ 
			PAN\cite{pan} &ResNet18& {89.3} & 81.0 & 85.0 & 39.6 \\
			TextDCT\cite{textdct} &ResNet50& {87.2} & 82.7 & {84.9} & 15.1 \\
			%			DB\cite{db}  &ResNet18& {88.3} & 77.9 & 82.8 & \textcolor{red}{\textbf{50}} \\
			ASTD \cite{astd}  &ResNet101  &85.4&81.2 &83.2 &- \\
			CRAFT \cite{craft} &VGG16 &87.6 &79.9 &83.6 &-\\
			OKR \cite{okr} &ResNet18 & 85.8 & {80.9} & 83.3 & 40.5 \\
			PAN++ \cite{pan++} &ResNet18 &89.9 &81.0 &85.3 &38.3\\
			NASK \cite{nask} &ResNet50 &85.6 &83.2 &84.4 &8.2\\
			DBNet\cite{db} &ResNet50 &{87.1} &{82.5} &84.7 &32\\
			DBNet++\cite{db++} &ResNet50 &{88.9} &{83.2} &86.0 &28\\
			LeafText \cite{leaftext} & ResNet18 &88.9 &83.2 &\textcolor{red}{\textbf{87.3}} &-\\
			LPAP\cite{lpap}  &ResNet50  &87.3 &79.8 &83.4 &-\\ \hline	
			%			DB++\cite{db++} &ResNet18& {87.4} & 79.6 & 83.3 & 48 \\
		\textbf{FEPE}  (Syn)  &ResNet18&90.8  & {79.5} & {84.8} & \textcolor{red}{\textbf{50}} \\
			\textbf{FEPE} (Syn)  &ResNet50& 91.3 & {81.9} & \textcolor{blue}{\textbf{86.4}} &32 \\
			
			\hline
		\end{tabular}
		\label{total}
	}
	
\end{table}
\subsubsection{Influence of the pre-training}
Pre-training using additional datasets has a significant impact on the detection results. The ICDAR2015 dataset is minimally affected by pre-training, resulting in only slight improvements of 0.2$\%$ and 0.5$\%$ after pre-training on SynthText and MLT, respectively. In contrast, the MSRA-TD500 dataset is the most affected by pre-training. After pre-training on SynthText, the performance improved by 2$\%$, while pre-training on MLT resulted in an F-measure improvement of 5.3$\%$. For the Total-Text dataset, pre-training on SynthText was more effective than pre-training on MLT, with performance improvements of 2.8$\%$ and 2$\%$, respectively. Pre-training on both datasets resulted in gains of 1.9$\%$ for CTW1500. Based on the experimental results, we can conclude that MLT yields more significant performance improvement for multi-directional text datasets, while SynthText is better suited for datasets that contain a large amount of irregular-shaped text.
\subsubsection{Visual comprasion}
We visualize and compare the detection performance of the proposed method with the baseline  (only predict text kernel maps and text maps), as seen in Fig. \ref{first}. The baseline model misidentifies an instance as two text instances when confronted with texts containing large gaps (Fig. \ref{first} first line). The proposed FEM improves the cohesiveness of pixels within the same text instance, thus exhibiting a better performance in this particular case. Moreover, the baseline model has a major drawback of misclassifying certain patterns that have similar textures with text. This issue is alleviated by the PEM, which confirms whether a given pixel belongs to text by emphasizing its perceptual surroundings.
\begin{figure*}
	\centering
	\includegraphics[width=0.88\linewidth,scale=1.0]{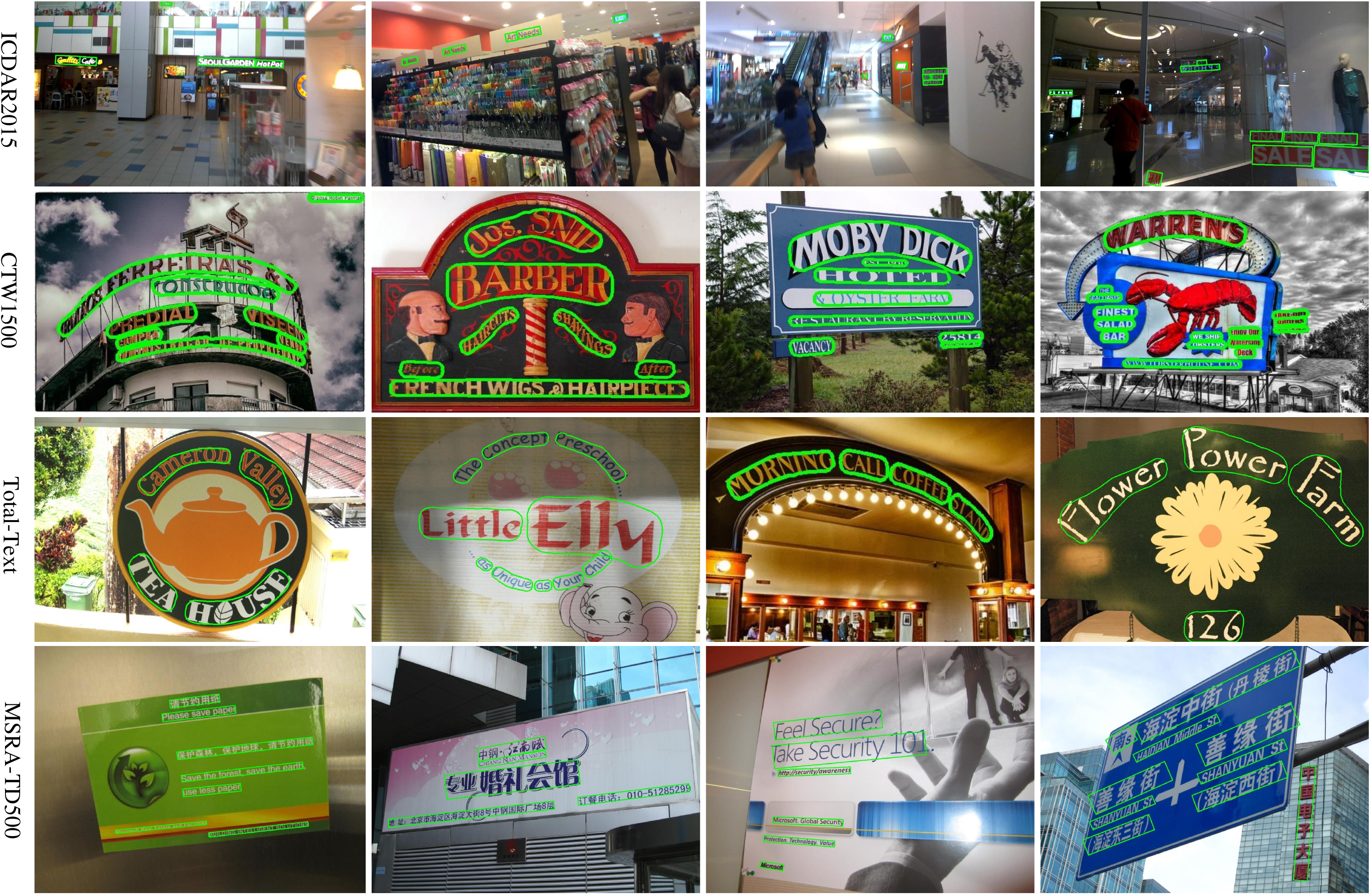}
	\caption{
		Visualizations of various types of text detection results are presented, including horizontal text, rotated text, and irregular text. The first and second rows of samples are from ICDAR2015 and CTW1500, respectively, while the last two are from Total-Text and MSRA-TD500. The proposed method is able to handle text instances of arbitrary shapes effectively.}\label{img_visual}
\end{figure*}

\indent
\begin{figure*}[t]
	\centering
	
	\subfigure[vs. FCENet \cite{fcenet}]{
		\begin{minipage}[t]{0.20\linewidth}
			
			\includegraphics[width=1\linewidth,height=8.5cm]{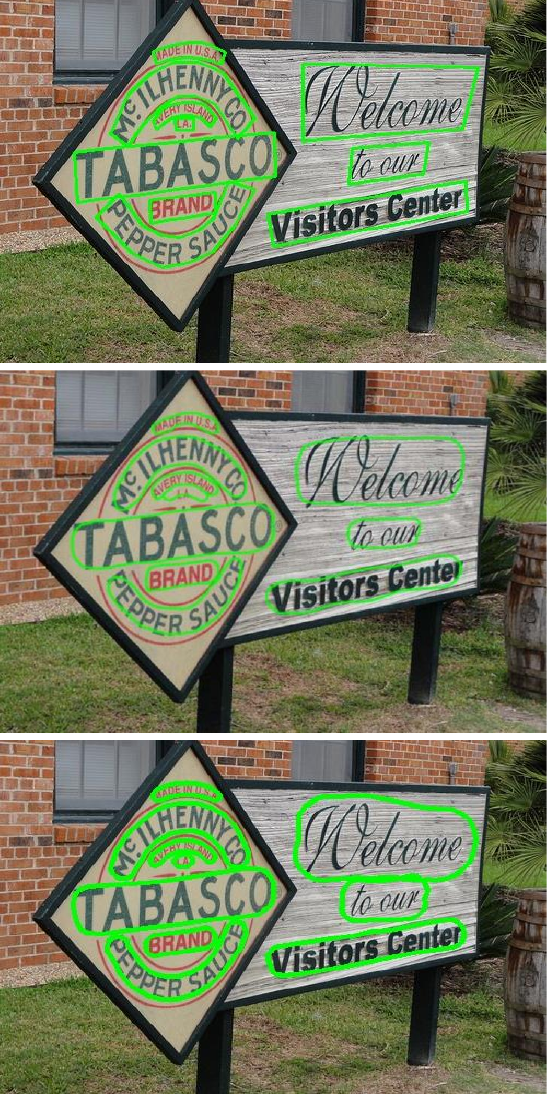}
			%\caption{fig1}
		\end{minipage}%
		
	}\label{com1a}
	\hspace{-3mm} 
	\subfigure[vs. TextRay\cite{textray}]{
		\begin{minipage}[t]{0.245\linewidth}
			
			\includegraphics[width=1\linewidth,height=8.5cm]{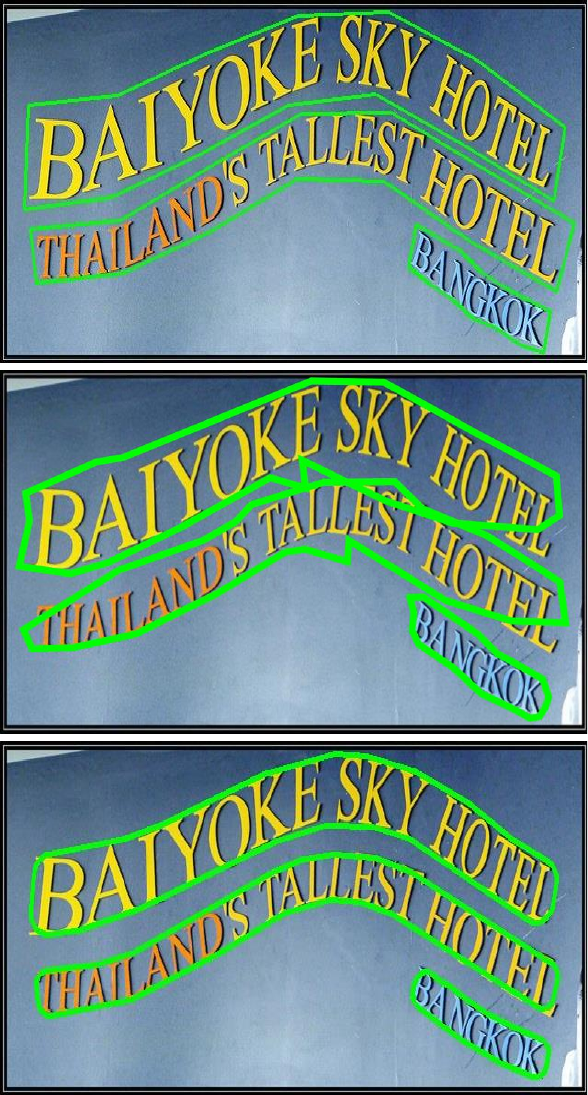}
			%\caption{fig1}
		\end{minipage}%
		
	}\label{com1b}
	\hspace{-3mm}
	\subfigure[vs. PAN \cite{pan}]{
		\begin{minipage}[t]{0.20\linewidth}
			
			\includegraphics[width=1\linewidth,height=8.5cm]{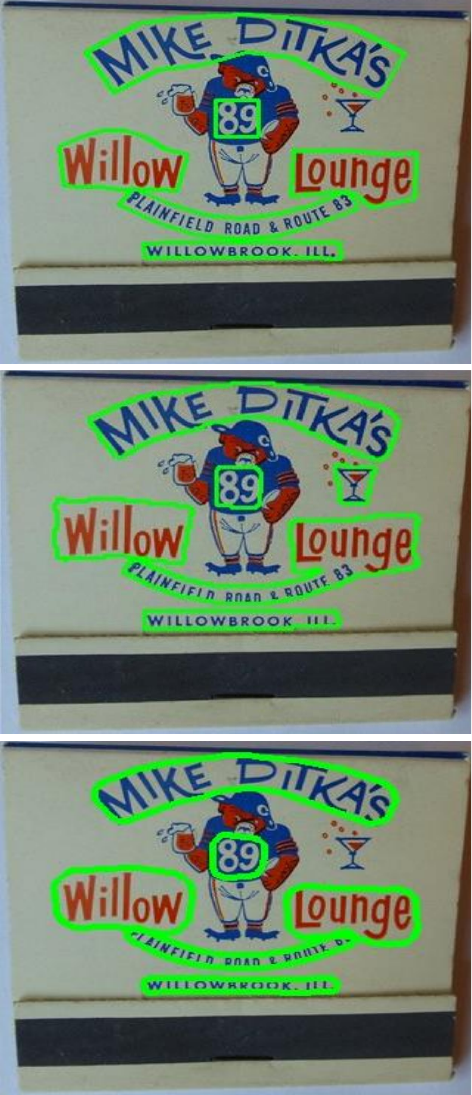}
			%\caption{fig1}
		\end{minipage}%
		
	}\label{com1c}
	\hspace{-3mm}
	\subfigure[vs. LPAP \cite{lpap}]{
		\begin{minipage}[t]{0.20\linewidth}
			
			\includegraphics[width=1\linewidth,height=8.5cm]{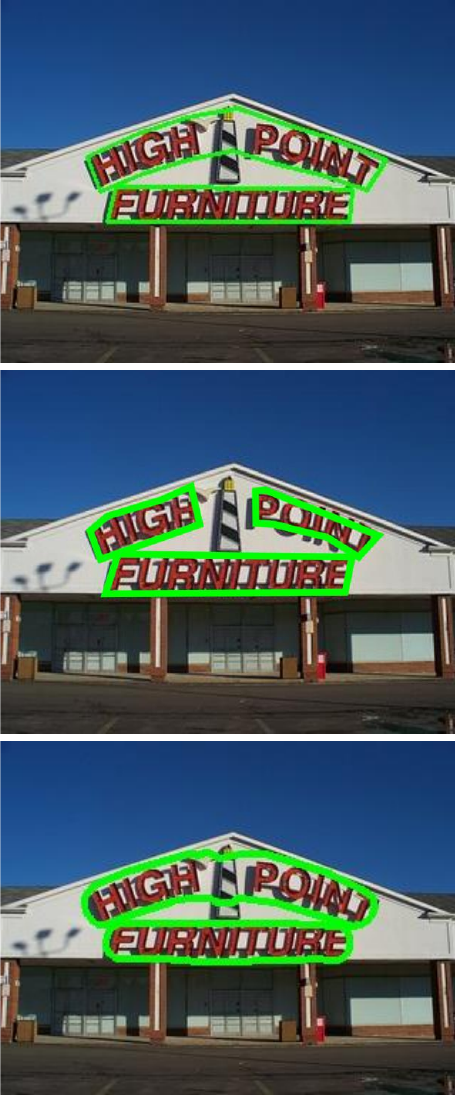}
			%\caption{fig1}
		\end{minipage}%
		
	}\label{com1d}
	
	\caption{ Some visual comparisons with FCENet \cite{fcenet}, TextRay\cite{textray}, PAN \cite{pan}, and LPAP \cite{lpap}. The middle row displays the results for FCENet \cite{fcenet}, PAN \cite{pan}, DBNet \cite{db}, and DBNet++ \cite{db++}, respectively. The top row is the labels corresponding to the images. The bottom row displays our detection results. }
	\label{compare}
\end{figure*}

\subsection{Comparison with State-of-the-Art Methods}     
The comparison with SOTA approaches is built on four benchmarks containing various text types. ICDAR2015 and MSRA-TD500 are word-level and line-level multi-directional text datasets. Total-Text and CTW1500 are word-level and line-level irregular-shaped text datasets. The advantages of FEPE are further analyzed through comparison.

\textbf{Evaluation on MSRA-TD500.}
%It is a multi-language and multi-directional long text dataset. We resize the short side to 736 while keeping ratio during inference stage.
%
%As exhibited in Table. \ref{td500}, the proposed FEPE achieves 85.6$\%$ (pretrained on SynthText) and 89.3$\%$ (pretrained on MLT) on F-measure when adopted ResNet18 as backbone. Moreover, the method is equipped with ResNet50 bring 2.1$\%$ (pretrained on SynthText) improvement.  Our approach is substantially superior to the existing SOAT methods in terms of performance and speed. Benfiting form FEM strength the identification of feature between instances at various scales, FEPE surpasses ADNet 0.3$\%$ on F-measure which not mention the FPS. Comparied with DBNet++, our approach has improved in both speed and performance. 
%
%We can see from Fig. \ref{img_visual} and Table \ref{td500} that the superority of FEPE for detecting multi-directional long text instances. 
MSRA-TD500 is a  multi-directional line-level labeled text dataset that contains Chinese and English. During the inference stage, we resize the short side of the input images to 736. As we can see from Table. \ref{td500}, the proposed FEPE achieves 86.0$\%$ (pre-trained on SynthText) and 89.5$\%$ (pre-trained on MLT) for F-measure when ResNet18 is used as backbone. Moreover, the method equipped with ResNet50 brings a 2.1$\%$ (pre-trained on SynthText) improvement. The proposed approach significantly outperforms existing SOTA methods regarding both performance and speed. Benefiting from the FEM that strengthens the identification of features between instances at various scales, FEPE surpasses ADNet \cite{ad} by 0.3$\%$ on F-measure. Compared with DBNet++ \cite{db++}, our approach has improved both speed and performance. As seen from Fig. \ref{img_visual} and Table \ref{td500}, FEPE is particularly effective for detecting multi-directional long text instances.

\begin{table}[t]
	
	\center
	\renewcommand\arraystretch{1.0}
	\setlength{\tabcolsep}{2mm}
	{
		\caption{  Comparison with existing advanced approaches on the ICDAR2015.  ``\textcolor{red}{\textbf{Red}}'' and ``\textcolor{blue}{\textbf{Blue}}'' represent the optimal and sub-optimal performance, respectively.}
		\begin{tabular}{cccccc}

			\hline Method &Backbone  & P & R & F & FPS \\
			\hline 
			
			%CTPN  & $74.2$ & $51.6$ & $60.9$ & $7.1$ \\
			EAST\cite{east}  &VGG16 & 83.6 & 73.5 & 78.2 & 13.2 \\
			PixelLink\cite{pixellink} &VGG16  &85.5 &82.0 &83.7 &- \\
			%TextBoxes++\cite{liao2018textboxes++} & $87.2$ & $76.7$ & $81.7$ & $11.6$ \\
			PSE-1s\cite{pse} &ResNet50 & 86.9 & {84.5} & 85.7 & 1.6 \\
			TextSnake\cite{textsnake} &VGG16 & 84.9 & 80.4 & 82.6 & 1.1\\
			
			Boundary\cite{wang2020all} &ResNet50 & 88.1 & 82.2 & 85.0 & - \\
			FCENet\cite{fcenet} &ResNet50& 90.1 & 82.6 & 86.2 &- \\
			%		SAE\cite{tian2019learning}  & $85.1$ & $84.5$ & $84.8$ & $3.0$ \\
			KPN\cite{9732893} &ResNet50& 88.3 & {88.3} & 86.5 & 6.3 \\
			DBNet++ \cite{db++}&ResNet50  &{90.9} & 83.9 & \textcolor{blue}{\textbf{87.3}} & {10} \\
			DBNet++ \cite{db++}&ResNet18  &{90.1} & 77.2 & 83.1 & {44} \\
			DBNet \cite{db} &ResNet50 & {91.8} & 83.2 & \textcolor{blue}{\textbf{87.3}} & {12} \\ 
			LOMO \cite{lomo} &ResNet50 & {91.3} & 83.5 & 87.2 & - \\
			%			LSFL(Res50)(1152) & $91.1$ & ${83.9}$ & $\mathbf{87.4}$ & $\mathbf{12}$ \\
			CM-Net\cite{cm} &ResNet18& 86.7 & 81.3 & 83.9 &34.5 \\ 
			PAN\cite{pan} &ResNet18 & 84.0 & {81.9} & 82.9 & 26.1 \\
			ZTD\cite{zoom} &ResNet18 &87.5 &79.0 &83.0 &\textcolor{red}{\textbf{48.3}}\\
			Spotter\cite{8812908} &ResNet50& 85.8 & {81.2} & 83.4 & 4.8 \\
			BiP-Net \cite{bip} &ResNet18 & 86.9 & {82.1} & 83.9 & 24.8 \\
			PAN++ \cite{pan++} &ResNet50 &91.4 &83.9 &\textcolor{red}{\textbf{87.5}} &12.6 \\ 
			ASTD \cite{astd}  &ResNet101  &88.8&82.6 &85.6 &- \\
			LeafText \cite{leaftext} &ResNet50 &88.9 &82.3 &86.1 &- \\
			LPAP\cite{lpap}  &ResNet50  &88.7 &84.4 &86.5 &-\\ \hline
			\textbf{FEPE} (Syn)  &ResNet18& 87.3 & {79.4} & {83.2} & \textcolor{blue}{\textbf{48}} \\
			%			FEPE  &ResNet18& 88.6 & {78.9} & {83.5} & \textcolor{red}{\textbf{48}} \\
			\textbf{FEPE} (Syn)  &ResNet50 & 89.8 & {84.9} & \textcolor{blue}{\textbf{87.3}} & 12 \\
			\hline
		\end{tabular}
		\label{ic15}
	}
	
\end{table}

\textbf{Evaluation on Total-Text and CTW1500.}
The Total-Text and CTW1500 datasets contain lots of varying shape and orientation texts. During the testing stage, the short side of input images is resized to 800. As shown in Table \ref{total}, our approach outperforms LPAP \cite{lpap}, ASTD \cite{astd}, and TextDCT \cite{textdct} by 3.0$\%$, 3.2$\%$, and 1.5$\%$, respectively, while also maintaining a faster speed. The proposed PEM helps FEPE in perceiving the environment around a pixel to confirm whether the pixel is text or not. Even though LeafText \cite{leaftext} is superior to ours, it needs complex post-processing and not mentation the speed, which limits it to apply in the real world significantly. Moreover, when adopting ResNet18 as the backbone, our method achieves competitive performance while maintaining fast speed. Unlike Total-Text, CTW1500 is a line-level annotated irregular text dataset. As shown in Table \ref{td500}, FEPE achieves the F-measure of 85.5 $\%$ and 86.0$\%$ when adopting ResNet18 and ResNet50 as the backbone, which surpassing the existing SOTA method DBNet++ \cite{db++} by 2.9$\%$ and 0.7$\%$, respectively, while maintaining a speed advantage. Even our approach using ResNet18 as the backbone is still superior to some existing SOTA methods using ResNet50. This further demonstrates the superiority of FEPE. We visualize some samples from Total-Text and CTW1500 in Fig. \ref{img_visual} to demonstrate the effectiveness of FEPE. 
Furthermore, in Fig.  \ref{compare}, we compare the visible results of our approach to SOTA methods. TextRay \cite{textray} is a regression method that fails to fit instances with particularly uneven aspect ratios accurately. FCENet\cite{fcenet} and PAN\cite{pan}, which focus only on pixel information, incorrectly classify some patterns similar to text as text and have problems with adjacent text sticking in PAN \cite{pan}. Additionally, LPAP \cite{lpap} misclassifies one text as two instances. Since FEPE focuses on instance-level features, it effectively addresses these problems and showcases the superiority of the proposed method.

\textbf{Evaluation on ICDAR2015.} This dataset contains images with complex backgrounds, low resolution, and dim lighting, making scene text detection challenging. The large variation in scale and multiple orientations are additional reasons for the difficulty in detecting instances. As shown in Table \ref{ic15}, when adopting ResNet50 as the backbone and resizing the short side to 1152, the proposed method achieves 89.8$\%$, 84.9$\%$, and 87.3$\%$ on precision, recall, and F-measure, respectively. The proposed FEPE surpasses the existing SOTA method LeafText \cite{leaftext} by 1.2$\%$ in terms of F-measure, even though LeafText uses a complex post-process without mentioning the speed. Moreover, the proposed method outperforms most existing SOTA approaches (such as KPN \cite{9732893},  LPAP \cite{lpap}, and FCENet \cite{fcenet}) on performance and speed. Although DBNet++ \cite{db++} achieves the same performance, mainly because it introduces an extra attention module that sacrifices speed. PAN++ \cite{pan++} surpasses ours 0.2$\%$ in F-measure, which is mainly because it uses ICDAR2017-MLT to pre-train, but the proposed method uses SynthText.  Using real datasets to pre-train generates better results than synthetic datasets. The objective evaluation metrics presented in Table \ref{ic15} and the visualization results in Fig. \ref{img_visual} effectively demonstrate that our method can cope with multi-directional texts.

\subsection{Cross Dataset Text Detection}  
To show the shape robustness of the FEPE, we train it on one dataset and test it on another. Note that cross-train-test experiments adopt ResNet18 as the backbone. We divided the four datasets used in the experiments into two categories based on the annotation style (word-level or line-level). As shown in Tab. \ref{tab_jiaocha}, FEPE achieves 71.4$\%$ and 77.3$\%$  of F-measure when training on Total-Text and ICDAR2015 and testing on ICDAR2015 and Total-Text. Compared with the  SOTA method CM-Net \cite{cm}, the proposed FEPE surpasses  1.7$\%$ and 5.2$\%$ in terms of F-measure, which shows the generalization ability on word-level texts. On the line-level datasets, FEPE achieves 79.7$\%$ and 80.7$\%$ when training on MSRA-TD500 and CTW1500 and testing on CTW1500 and MSRA-TD500. It is also substantially superior to TextField \cite{textfield}. Compared to the CM-Net \cite{cm}, the FEPE substantially outperformed that test on CTW1500. It is slightly inferior to the test on  MSRA-TD500. These experiments demonstrate that FEPE has excellent generalization for data of different shapes and that its data requirements are low compared to other methods. 
\begin{table}
	
	\center
	{
		\caption{  Two groups (word-level and line-level) cross-dataset evaluations, where IC15, Total, TD500, and CTW represent ICDAR2015, Total-Text, MSRA-TD500 and CTW1500 datasets, respectively. }
		\begin{tabular}{c|c|c|ccc}
			\hline
			Training &Testing &Methods &P &R & F \\ \hline
			\multirow{3}{*}{IC15}  &\multirow{3}{*}{Total} 
			&Textfield\cite{textfield} &61.5 &65.2 &63.3\\
			&&CM-Net \cite{cm}  & 75.8 & {64.5} &69.7 \\ 
			&&FEPE(ours) & 81.4 &63.5 & 71.4 \\ \hline 
			\multirow{3}{*}{Total} & \multirow{3}{*}{IC15} 
			&Textfield\cite{textfield} &77.1 &66.0 &71.1\\
			&&CM-Net \cite{cm}  & 76.5 & {68.1} &72.1 \\ 
			&&FEPE(ours) & 82.9 &72.5 & 77.3 \\ \hline
			\multirow{3}{*}{TD500} & \multirow{3}{*}{CTW} 
			&Textfield\cite{textfield} &75.3 &70.0 &72.6\\
			&&CM-Net \cite{cm}  & 77.2 & {69.7} &72.8 \\ 
			&&FEPE(ours) & 85.3 &74.8 & 79.7 \\ \hline
			\multirow{3}{*}{CTW} & \multirow{3}{*}{TD500} 
			&Textfield\cite{textfield} &85.3 &75.8 &80.3\\
			&&CM-Net \cite{cm}  & 85.8 & {77.1} &81.2 \\ 
			&&FEPE(ours) & 85.5 &86.3 & 80.7 \\ \hline
			
		\end{tabular}
		\label{tab_jiaocha}
	}
\end{table}
\subsection{Limitations} 

\begin{figure*}[t]
	\centering
	{
		\subfigure[Large amount of vertical texts.]{
			\begin{minipage}[t]{0.24\linewidth}
				\includegraphics[width=4.0cm,height=2.5cm]{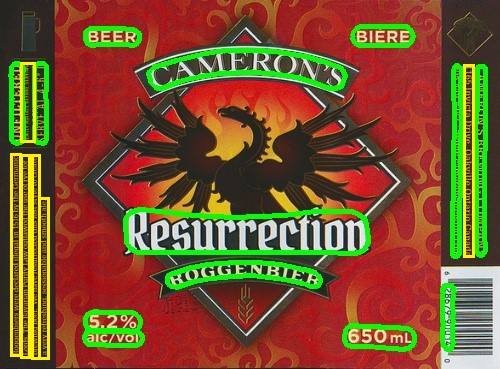}
				%\caption{fig1}
			\end{minipage}%
			\begin{minipage}[t]{0.24\linewidth}
				
				\includegraphics[width=4.0cm,height=2.5cm]{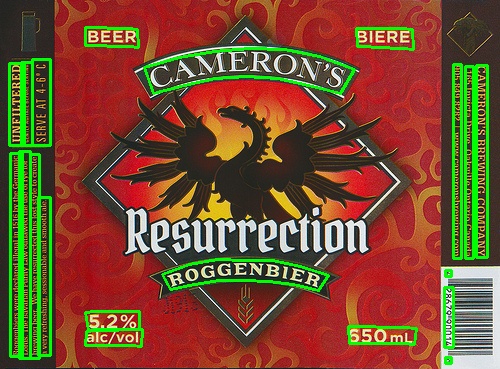}
				%\caption{fig1}
			\end{minipage}%
			
		}\label{lim_a}
%		\hspace{-3mm}
		\subfigure[The long text is separated by a barrier in the middle.]{
			\begin{minipage}[t]{0.24\linewidth}
				
				\includegraphics[width=4.0cm,height=2.5cm]{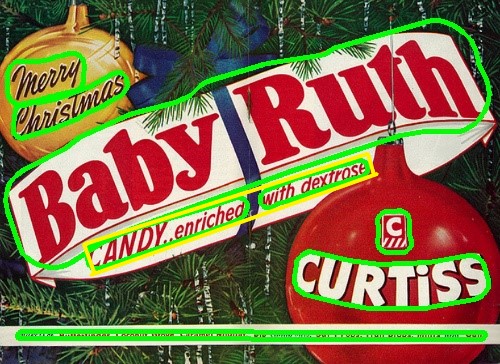}
				%\caption{fig1}
			\end{minipage}%
			\begin{minipage}[t]{0.24\linewidth}
				
				\includegraphics[width=4.0cm,height=2.5cm]{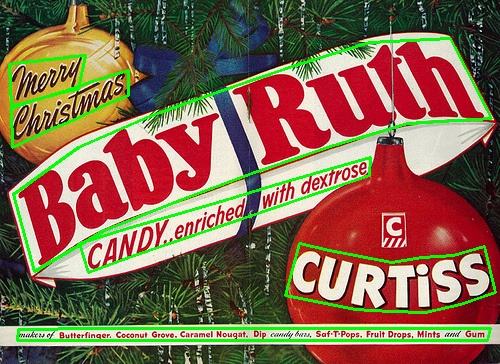}
				%\caption{fig1}
			\end{minipage}%
			
		}\label{lim_b}}
	\centering
	{
		\subfigure[Patterns that are similar to the texture of the text.]{
			\begin{minipage}[t]{0.24\linewidth}
				\includegraphics[width=4.0cm,height=2.5cm]{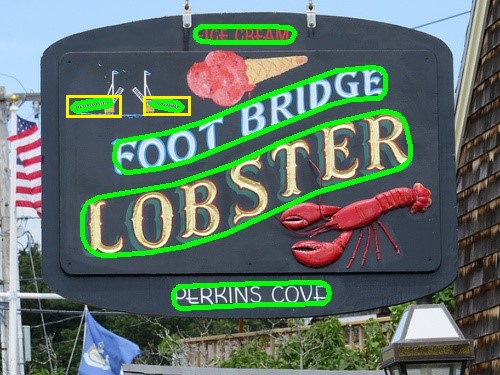}
				%\caption{fig1}
			\end{minipage}%
			\begin{minipage}[t]{0.24\linewidth}
				
				\includegraphics[width=4.0cm,height=2.5cm]{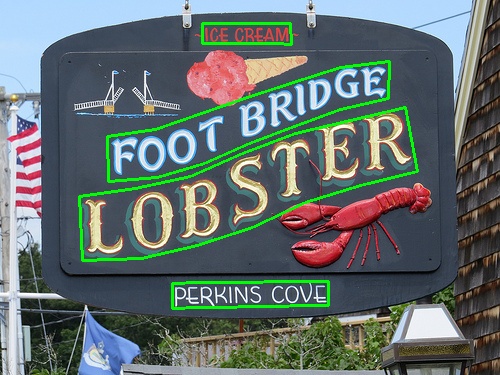}
				%\caption{fig1}
			\end{minipage}%
			
		}\label{lim_c}
%		\hspace{-3mm}
		\subfigure[Text with different colors for the same instance character.]{
			\begin{minipage}[t]{0.24\linewidth}
				
				\includegraphics[width=4.0cm,height=2.5cm]{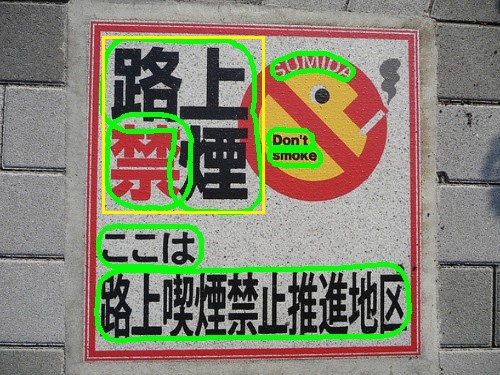}
				%\caption{fig1}
			\end{minipage}%
			\begin{minipage}[t]{0.24\linewidth}
				
				\includegraphics[width=4.0cm,height=2.5cm]{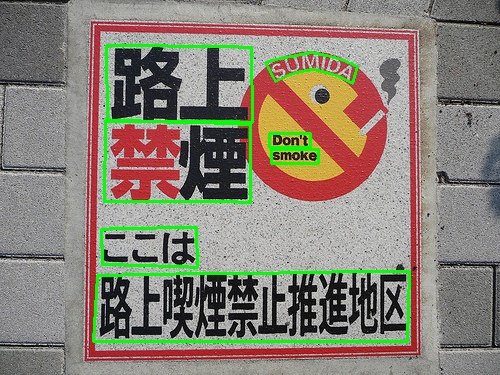}
				%\caption{fig1}
			\end{minipage}%
			
		}\label{lim_d}}
	
	\caption{Some limitations and drawbacks of the proposed FEPE include inadequate detection of vertical text, misclassification of the gaps in the long text as negative samples, and difficulties in accurately detecting text in texture and color interference. The left is our prediction in each group image, and the right is the corresponding ground truth. The incorrections in our results are used in yellow to mark.}
	\label{img_lim}
\end{figure*}
We demonstrate the superiority of the proposed FEM for instance-level feature extraction and the effectiveness of PEM for sensing the surrounding environment through ablation experiments. Also, the excellent performance on various datasets proves the advancedness of FEPE. In this phase, we further analyze the shortcomings and limitations of the FEPE. As shown in Fig. \ref{img_lim}(a), four typical errors are selected for detailed analysis. The lack of FEPE's ability to detect vertical text in the figure is mainly because vertical text instances are too rare in the training set and even in life. The model does not have enough samples to learn the features of vertical text, which can be considered to compensate for this shortcoming in the subsequent dataset construction. The long text in Fig. \ref{img_lim}(b) is truncated by an obstacle, resulting in one instance being misclassified by the model as two instances. This is the bottom-up approach of the segmentation method, which over-focuses on the underlying features and has an insufficient grasp of the holistic features of the instances. We can see in Fig. \ref{img_lim}(c) that some patterns similar to the text texture are misclassified as text. As we can see from Fig. \ref{img_lim}(d), the characters belonging to the same instance are wrongly divided into different instances due to different colors.
On the contrary, characters of different text instances are classified as the same instance due to the same color. These problems arise mainly because the model focuses only on visual information, for the language features are unaware. It is our future work to alleviate these problems.
\section{Conclusion}
\label{conclusion}
%\cite{*} 
In this paper, an arbitrary-shaped scene text detector is proposed that consists of FEM and PEM. The former encourages the model to distinguish instances of different scales, enhance the sense of belonging of pixels to their respective instances, and increase the cohesion of pixels belonging to the same sample. The latter perceives the distribution of positive samples around each pixel to confirm whether the current pixel is a positive sample. The FEPE differs from existing segmentation-based methods which typically focus only on pixel-level information.
The proposed FEM and PEM enable the model to learn instance-level and region-level information, thus partially compensating for the insufficient global information extraction of bottom-up methods. Extensive experiments prove that the proposed FEPE significantly surpasses existing SOTA methods on four public benchmarks. We will continue to explore the relationships of multi-level information of scene texts to structure text knowledge systems in the future.

%
%
%
%{\appendix[Proof of the Zonklar Equations]
%Use $\backslash${\tt{appendix}} if you have a single appendix:
%Do not use $\backslash${\tt{section}} anymore after $\backslash${\tt{appendix}}, only $\backslash${\tt{section*}}.
%If you have multiple appendixes use $\backslash${\tt{appendices}} then use $\backslash${\tt{section}} to start each appendix.
%You must declare a $\backslash${\tt{section}} before using any $\backslash${\tt{subsection}} or using $\backslash${\tt{label}} ($\backslash${\tt{appendices}} by itself
% starts a section numbered zero.)}

\ifCLASSOPTIONcaptionsoff
\newpage
\fi
%\linespread{1}

\bibliographystyle{IEEETran}
\bibliography{IEEEabrv}

\end{document}